\documentclass{article}

\usepackage{arxiv}
\usepackage{authblk}

\usepackage[style=apa,natbib=true]{biblatex}
\usepackage{algorithm}
\usepackage{algpseudocode}
\newlength\myindent
\newcommand\mydots{\makebox[1em][c]{.\hfil.\hfil.}}

\usepackage[export]{adjustbox}[2011/08/13]
\usepackage{hyperref}
\hypersetup{hidelinks,
colorlinks=true,linkcolor=ourblue,
urlcolor=ourblue,citecolor=ourviolet,
bookmarksopen=false}
\usepackage{xcolor}
\usepackage{ifthen}

\usepackage{amsfonts}

\PassOptionsToPackage{fleqn}{amsmath} 
\usepackage{amsmath}
\usepackage{comment}

\usepackage{xspace}
\makeatletter
\DeclareRobustCommand\onedot{\futurelet\@let@token\@onedot}
\def\@onedot{\ifx\@let@token.\else.\null\fi\xspace}

\def\ie{i.e\onedot}

\makeatother

\usepackage{xcolor}
\definecolor{ourblue}{rgb}{0.368,0.507,0.71}
\definecolor{ourgreen}{rgb}{0.56,0.692,0.195}
\definecolor{ourred}{rgb}{0.923,0.386,0.209}
\definecolor{ourviolet}{RGB}{59, 58, 126}
\definecolor{ourorange}{RGB}{232, 122, 18}

\newcommand{\fs}[1]{{#1}} 

\newcommand*{\figuretitle}[1]{%
    {\centering
    \textbf{#1}
    \par\medskip}
}

\title{When to be critical? \\Performance and evolvability in different regimes of neural Ising agents.}

\author[1, 2]{
 Sina Khajehabdollahi$^\star$ (\texttt{@sinak12345})
}
\author[1, 2]{
 Jan Prosi$^\star$ (\texttt{@Jan\_psi})
}
\author[1, 2]{
 Emmanouil Giannakakis (\texttt{@m\_giannakakis})
}
\author[3]{
 Georg Martius (\texttt{@GMartius})
}
\author[1, 2, 4]{
 Anna Levina (\texttt{@SelfOrgAnna})
}

\affil[1]{Department of Computer Science, University of T\"ubingen, T\"ubingen, Germany}
\affil[2]{Max Planck Institute for Biological Cybernetics, T\"ubingen, Germany }
\affil[3]{Max Planck Institute for Intelligent Systems, T\"ubingen, Germany}
\affil[4]{Bernstein Center for Computational Neuroscience  T\"ubingen, T\"ubingen, Germany}
\affil[ ]{$^\star$These authors contributed equally to the work.}

\begin{document}
\maketitle

\begin{abstract}
It has long been hypothesized that operating close to the critical state is beneficial for natural, artificial and their evolutionary systems.
We put this hypothesis to test in a system of evolving foraging agents controlled by neural networks that can adapt agents' dynamical regime throughout evolution.
Surprisingly, we find that all populations that discover solutions, evolve to be subcritical.
By a resilience analysis, we find that there are still benefits of starting the evolution in the critical regime. Namely, initially critical agents maintain their fitness level under environmental changes (for example, in the lifespan) and degrade gracefully when their genome is perturbed.
At the same time, initially subcritical agents, even when evolved to the same fitness, are often inadequate to withstand the changes in the lifespan and degrade catastrophically with genetic perturbations. Furthermore, we find the optimal distance to criticality depends on the task complexity. To test it we introduce a hard and simple task: for the hard task, agents evolve closer to criticality whereas more subcritical solutions are found for the simple task.
We verify that our results are independent of the selected evolutionary mechanisms by testing them on two principally different approaches: a genetic algorithm and an evolutionary strategy. 
In summary, our study suggests that although optimal behaviour in the simple task is obtained in a subcritical regime, initializing near criticality is important to be efficient at finding optimal solutions for new tasks of unknown complexity. 
\end{abstract}


\section{Introduction}
Operating close to the critical point at a second order phase transition has long been associated with optimal performance of complex systems. Several biological systems, such as gene regulatory networks \citep{balleza2008critical, ramo2006perturbation}, neural networks \citep{tkacik2015thermodynamics, schneidman2006neural_cultures, Beggs2004}, collectively behaving cells \citep{halley2009collective_cells, depalo2017collective_cells}, swarms \citep{cavagna2010scale, chate2014insect}, or populations of co-evolving, communicating agents \citep{hidalgo2014information} have been shown to operate close to a critical point. 
 Criticality has been associated with an ability to solve complex tasks \citep{villegas2016noise_subcritical},  optimal information transmission and sensitivity~\citep{Kinouchi2006, Beggs2007, boedecker2012, bertschinger2004}, flexibility towards changes in the environment and good evolvability \citep{aldana2007robustness_evolvability} in complex living systems \citep{kauffman1993origins}. In these models, different variations of the transitions and scaling exponents are considered: in the branching model related to neuroscience, it is a transition between absorbing and active state; in the Ising-like models, a transition between ordered and disordered state. However, as long as the model presents a second-order phase transition, most results remain qualitatively unchanged.
 All these optimized properties provide an adaptive advantage in natural environments, leading to the assumption that evolutionary dynamics push biological systems close to the critical regime.

 On the other hand, it has been suggested that the ubiquitous presence of noise in nature pushes living systems into a more robust subcritical regime.
 For example, in an evolutionary model of Random Boolean Networks (RBNs), decreasing the system size, making the task less complex, or introducing noise to the system pushes the optimal regime further into the subcritical range \citep{villegas2016noise_subcritical}. Similarly \citet{pauli_ramo2007_critical_information_propagation_subcritical_noise} observed, that whereas information propagation is maximized in critical RBNs, the optimal regime shifts slightly into the subcritical regime under the presence of noise.
 In recordings from the nervous systems of different animals, slightly subcritical behaviour was observed~\citep{Priesemann2014, wilting2019between}. 
 A related phenomenon has been observed on neuromorphic chips, which optimally perform in simple tasks when in the subcritical regime, whereas harder tasks require progressively more critical dynamics \citep{cramer_control_2020}.
The escape dynamics of schooling fish remain in subcritical state even under pharmacological manipulation that increases alertness, though more alert fish self-organize closer to critical state~\citep{poel2021subcritical}.   
 Finally, for some applications the combination of systems operating at different distances from criticality has been shown to lead to optimal results \citep{zierenberg2020tailored}. The disordered supercritical state has been universally observed to perform poorly \citep{villegas2016noise_subcritical, kauffman1993origins}. There are different critical transitions in the examples mentioned above, thus different definitions of what it means to be subcritical. For example, for the order/disorder transition, subcritical refers to the ordered state. At the same time, for the branching network transition to an active state, subcritical means that perturbations are rapidly dying out. Interestingly, the optimal behaviour was found in the subcritical regime for both these transition types. 

\begin{figure}
\centering
\includegraphics[height=5.0in]{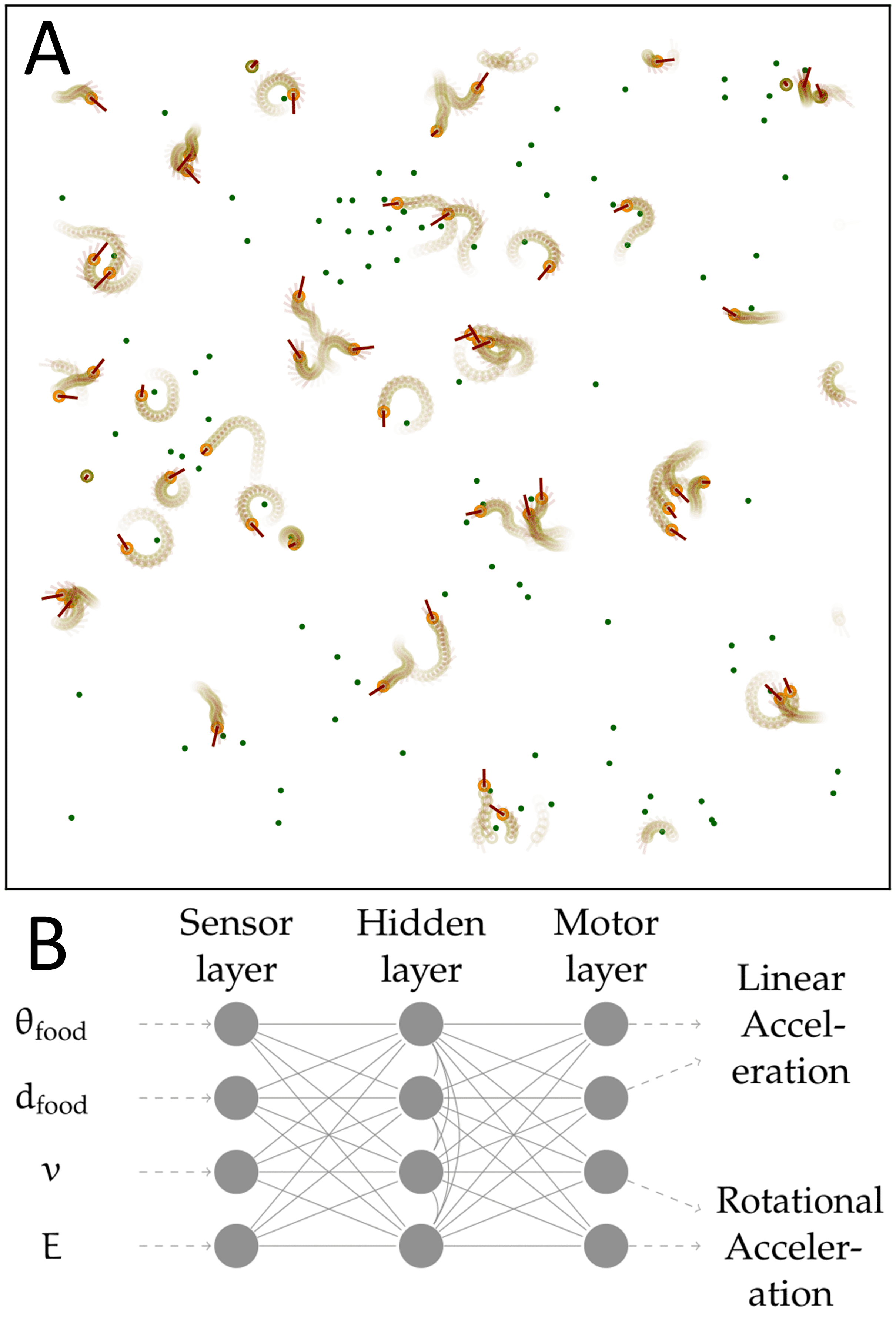}
\caption{
Snapshot of population dynamics and schematic representation of the control network. \textbf{A:}
Environment with 50 organisms (red circles with trails) foraging for food (green dots).
\textbf{B:} A network with $12$ neurons. Four sensory, four hidden and four motor neurons. All allowed edges are displayed. The exact topology and edge weights are subject to change by the evolutionary algorithm.
}
\label{fig:methods}
\end{figure}

The benefits of criticality for the evolvability of living systems have been associated with the genotype-phenotype coupling.
Specifically, it has been shown  \citep{dejong2006evolutionary_unified_book} that a tight genotype-phenotype coupling leads to optimal evolvability.
Due to this coupling, the dynamical regime has an impact on the properties of the fitness landscape. In an RBN model, the super- and subcritical regimes were shown to disturb the genotype-phenotype coupling \citep{kauffman1993origins} and lead to either very rugged or overly flat fitness landscapes.
A rugged fitness landscape means that the evolutionary dynamics are just a random search and thus inefficient in high dimensions~\citep{kauffman1987waiting_times_double}. On the other hand, 
a very flat landscape dampens the optimization process. Both phenomena result in a complexity catastrophe, where an increase of system size leads to a failure to discover satisfying solutions with evolutionary search.
Critical RBNs result in intermediately rugged fitness landscapes which allow for efficient hill climbing search and are less prone to the complexity catastrophe.

Our previous study~\citep{Prosi}, examined how the dynamical regime of populations of evolving organisms influences their ability to solve a task. Our investigation used a simple foraging game of scalable difficulty, where organisms can gain energy by eating food particles and consume energy when moving. We optimized the Ising networks controlling the organisms using a simple genetic algorithm which allowed us to analyze the changes in the dynamical regimes during evolution.
In addition, we proposed a potential answer to the question of which dynamical regime demonstrates the best performance and stability with respect to changes in the environment. Still, this study left unclear the extent to which the results are determined by the choice of evolutionary strategy or if they represent a general trend regardless of which EA is used for optimization.
Here, we conduct a more detailed analysis of the dynamics underlying our network model and extend the previous findings by comparing the behaviour of two distinct evolutionary strategies. Overall we  confirm that our results are not dependent on the exact algorithm used to train the model.  

\section{Methods}

\fs{We investigate a 2D environment where organisms controlled by individual neural networks forage for food}.
    Each organism gains energy by eating food particles and consumes energy by moving. The organisms eat the food particles by running over them and share their environment with other organisms in the same generation. This multi-agent environment is chosen to allow for the environment to complexify as agents evolve and become more adept to their task, thereby changing the distribution of input signals an individual experiences in its lifetime. Furthermore, this type of multi-agent environment forces the agents into a strategic competition with themselves, which again encourages a richer environment.
Motivated by the results in \citep{hidalgo2014information} where environment complexity and the optimal dynamical state were positively correlated, we introduce a mechanism to complexify our environment. We can increase the difficulty of the task by requiring the organism's velocity to be below a certain threshold when running over food in order to be able to consume it.
The fitness of an organism is determined by its average energy throughout its lifespan. We use two distinct evolutionary algorithms to optimize the network controlling the organisms and compare their performance.

\subsection{Organism}

The organisms in our model are controlled by an Ising neural network (INN) that has been previously used in \citep{aguilera2017ising_neural_network} as well as \citep{khajehabdollahi2020sinas_paper}.
The Ising network consists of $N$ neurons that can be in one of two states $ s_i \in \{-1, 1 \}, i = 1, \dots N$. All neurons are split in three classes: sensory neurons that only receive input from the sensors, motor neurons that control the agent, and hidden units used for computations.
Their connectivity is described by the adjacency matrix $A \in \{0,1\}^{N \times N}$ and the weight matrix $J \in [-2, 2]^{N \times N}$, as shown in Figure~\ref{fig:methods}B.
No connections between sensor and motor neurons are allowed by adjacency matrix at any time.
Following the Ising model, each network activation pattern (vector of states of all neurons) has an associated energy
\begin{eqnarray}
e(s_1,\ldots,s_N)= -\sum\limits_{i,j} A_{ij}J_{ij} s_i s_j.
\label{eq:ising_net_energy}
\end{eqnarray}
The network stochastically minimizes the energy by following Glauber dynamics:
At each network iteration, all non-sensor neurons are updated in a random order and the state of neuron $i$ changes from $s_i$ to $-s_i$ with probability:
\begin{eqnarray}
p_i &=& \frac{1}{1+e^{\beta \cdot \Delta e_i}},
    \label{eq:spin_flip}\\
\Delta e_i &=& e (s_1,\mydots,s_i,\mydots,s_N) - e (s_1,\mydots,-s_i,\mydots,s_N), \nonumber
\end{eqnarray}
where $\beta$ is the inverse temperature of the network ($\beta = 1 / (T \cdot k_B)$,  $k_B$ is the Boltzmann constant which we set to one and omit for simplicity) and $\Delta e_i$ is the change in the energy of the network that is caused by the spin-flip of the $i_{th}$ neuron (changing its state $s_i$ to $-s_i$).
The energy change $\Delta e_i$ is determined by the connectivity matrix $J$ and the states of neighboring neurons.
A decrease in energy (negative energy change) leads to a greater likelihood of a flip. The parameter $\beta > 0$ controls the likelihood of energetically unfavourable flips.
A large $\beta \gg 1$ leads to deterministic network behaviour dominated by the connectivity, whereas a smaller $\beta$ leads to more random behaviour. For each time step in the simulation, the motor neurons of an agent are read out and apply an action, and the sensory neurons are then updated. The model must then thermalize according to the new values of the sensor neurons by updating its state via Eq.~\ref{eq:spin_flip}. In principle, the number of iterations required to converge to equilibrium is a function of the connectivity matrix, the temperature of the model, and the distribution of sensor values. It is known that near the critical point it takes more time to reach an equilibrium state. However, for practical reasons, we fix the number of iterations to 10 thermalization steps. An analysis on the sensitivity of the model to the thermalization time is provided in appendix \ref{sec:app:thermal}.
In principle this hyper-parameter acts as the amount of time an agent has to ``think'' about its new sensory inputs, and may be biologically constrained.

An organism has four input neurons that receive information about the angle $\theta_\mathrm{food}$ and distance $d_\mathrm{food}$ from the closest food particle as well as its own velocity $v$ and energy $E$ (this energy is distinct from the Ising energy $e$).
Moreover, each organism has four output neurons that control linear and rotational acceleration (2 neurons each) and $N_h$ hidden neurons (Figure~\ref{fig:methods}B).
For each time step in the environment we assign a normalized real value to the sensor neurons according to the environmental input. The hidden and motor sensors can only obtain binary states (-1,1). We equilibrate the hidden and motor neurons for 10 iterations using a Metropolis algorithm \citep{metropolis1953metropolis_algorithm_original} which implements Eq.~\ref{eq:spin_flip} and subsequently read the states for the motor neurons. The agent accelerates in case both neurons of a motor unit are in agreement and have positive states, decelerates in case both are in agreement and have negative states, and does nothing if the neurons are in disagreement and have opposite states.

For most simulations we use a hidden layer with 4 neurons, ($N_h = 4$). Additionally,  in order to study whether our methods perform well with larger networks, we also simulate a network with $N_h = 20$.
At the beginning of each simulation, an organism is provided with an amount of initial energy $E_\mathrm{init}=2$.
Movement reduces energy and consuming food particles increases it.
We consider two versions of this environment:  In the \emph{simple task}
organisms consume food when passing over it.
In the \emph{hard task} organisms have to slow down and almost stop to be able to consume food.
Unless stated differently, a simulation lasts for a \emph{lifespan} of $t=2000$ time steps after which the evolutionary algorithm (EA) is applied, and the task is simple.
50 INN-controlled organisms are placed in a $2$D environment with periodic boundaries and ever-respawning food particles, conserved to a value of 100 (Figure \ref{fig:methods}A).

\subsection{Evolutionary algorithms}

\subsubsection{Genetic algorithm (GA)}

The genetic algorithm applied to the INNs consists of a combination of elitism, mutation, and mating.
At the end of the simulation described above, the fitness of each organism is defined as their mean energy throughout their lifespan. Subsequently, the 20 fittest organisms continue unchanged to the next generation.  15 more are added by duplicating the top 10 organisms with a 10\% chance of mutations. The remaining 15 are then populated by mating between the current population. The next generation therefore consists of 20 copied organisms, 15 possibly mutated, and 15 generated by mating. 
The mutation operation adds or deletes edges in $A$ (connections not present in Figure~\ref{fig:methods}B cannot be added), re-samples a random edge weight in $J$ from a uniform distribution $\mathcal{U}(-2, 2)$, and perturbs the inverse temperature with multiplicative Gaussian noise $\beta' = \beta \cdot \Delta \beta$, for $\Delta \beta \sim \mathcal{N}(1,\,0.02)$. Finally, the  mating operation randomly chooses two parents from the pool of the 35 individuals that either survived or were mutated duplicates, and  takes a weighted average of their connectivities $J$ and inverse temperatures $\beta$ to produce an offspring. In most of our simulations, the GA iterates for 4000 generations.

\subsubsection{Evolution strategy (ES)}\label{sec:ES}

We verify that our results are not contingent on the specific behaviour of the genetic algorithm described above, by employing 
an Evolution Strategy (ES) from the family of natural evolution strategies~\citep{wierstra2008natural, wierstra2014natural} with some modifications. 
In contrast to the GA these methods parametrize the population by a distribution over the genome and adapt its parameters.
The algorithm uses a multi-variate Gaussian distribution $\mathcal{N}(\mathbf{J}, \sigma \mathbb I)$ with mean $\mathbf{J}$ and fixed variance $\sigma$ (where $\mathbb I$ is the identity matrix). 
Using a fixed variance simplifies the algorithm and was reported to work well for neural network training \citep{salimans2017evolution}. 
The update of the mean $\mathbf{J}$ follows a gradient ascent on the fitness, estimated based on the fitness of $n$ sampled individuals. 
However, to make the gradient invariant to monotonous fitness transformations, a rank-based fitness is used, as in \citep{wierstra2014natural}. 
We start from the implementation by \citep{najarro2020meta} but
add elitism in two ways to the algorithm. 
We compute the gradient with respect to the best individual
 and keep a small fraction of elite individuals for the next generation. 
The update of the mean $\mathbf{J}$ is given by:
\begin{eqnarray}
    \label{eq:NES_update}
    \mathbf{J}_{t + 1} = \mathbf{J}^{\star}_t + \frac{\alpha}{n \sigma} \sum_{i=1}^{n} F \left ( \mathbf{J}_t + \sigma \epsilon_i \right ) \cdot \epsilon'_i,
\end{eqnarray}
where $\alpha$ is the learning rate, $\sigma$ is the standard deviation of our Gaussian search distribution, $n$ is the number of individuals generated (in our case equal to the population of the environment), $F \left (  \cdot \right ) $ the ranked fitness (our foraging task), $\epsilon_i = \mathcal{N}(0,\,1) $ is a Gaussian random vector, $J^\star$ is the parameter vector of the best individual, and $\epsilon'_i$ is the random vector $\epsilon_i$ relative to the best, \ie $\epsilon'_i= J_i + \sigma\epsilon_i - J_i^\star$. The values of $F$ are computed by first ranking all fitness values and then normalizing those ranks by subtracting their mean and dividing by their standard deviation.

We also update the inverse temperature ($\beta$) of the model in the same way, however we let this parameter evolve slower by setting its $\sigma$ to $\sigma_{\beta} = 0.1 \cdot \sigma$ in order to keep the ES algorithm comparable to the GA. Further motivation for this choice is that $\beta$ is a global parameter which should change slowly relative to the connectivity parameters. 
Furthermore, we do not include any decay rates in the learning rate and standard deviation in order to avoid conflating the dynamics of a decaying learning rule with any convergences that might occur due to selection pressures. 

Since each individual per generation is actually competing for the same resources as other individual (as opposed to running in an independent, parallel simulation), we also employ the use of elitism per generation to ensure that (6/50) previous well performing individuals are part of the next generation (the $\epsilon$s are computed accordingly). 
Furthermore, to allow for a sparse change in parameters during search we  set $50\%$ of entries in $\epsilon$ to zero. This parameter was introduced as a variable to control the genetic diversity between generations, again motivated by the idea that individuals are actually competing against one another and not running independently, and therefore should have to compete against similarly evolved individuals.

As commonly known, ES is sensitive to the hyper-parameter $\sigma$. For example in the works of \citet{Sehnke2010:PGPE} there is considerable work done to ensure that the $\sigma$s of different parameters are adaptive, where ensuring that the initial $\sigma_0$ is large enough to find a solution.
When chosen appropriately we found that our version of ES is optimizing the fitness of populations to comparable values as the GA. 
\renewcommand{\thealgorithm}{}
\begin{algorithm}
\begin{algorithmic}[1]
\caption{Evolution of Agents}\label{alg:model}
\For{$\text{generation} = 1 \textbf{ to } \text{Generations}$}
\State {\color{ourred} \textbf{Foraging Game in $2$D environment}}
\hspace{\algorithmicindent}\For{$t = 1 \textbf{ to } \text{organism's lifespan}$}
\State $\text{update sensor neurons}(\text{organism})$
\hspace{\algorithmicindent}\hspace{\algorithmicindent}\For{$\text{iter} = 1 \textbf{ to } \text{network iterations}$}
\Comment{\textit{Glauber update of INN}}
\hspace{\algorithmicindent}\hspace{\algorithmicindent}\hspace{\algorithmicindent}\For{$\text{non-sensor neuron} \textbf{ in } \text{INN}$}
\State $\text{potential spin-flip(neuron)}$ 
\EndFor
\EndFor
\State $\text{read motor neurons}(\text{organism})$
\State $\text{move in 2D environment}(\text{organism})$
\EndFor
\State {\color{ourred}\textbf{$\text{Evolve}(\text{population of organisms})$}} \Comment{\textit{Evolves $J$, $A$, and $\beta$}}
\State \qquad \text{using Genetic Algorithm or Evolution Strategy}
\State $\text{Reset 2D environment}$
\EndFor
\end{algorithmic}
\end{algorithm}

\subsection{Defining the dynamical regime of an organism}
\begin{figure}
\centering
\makebox[\linewidth]{
\includegraphics[width=\linewidth]{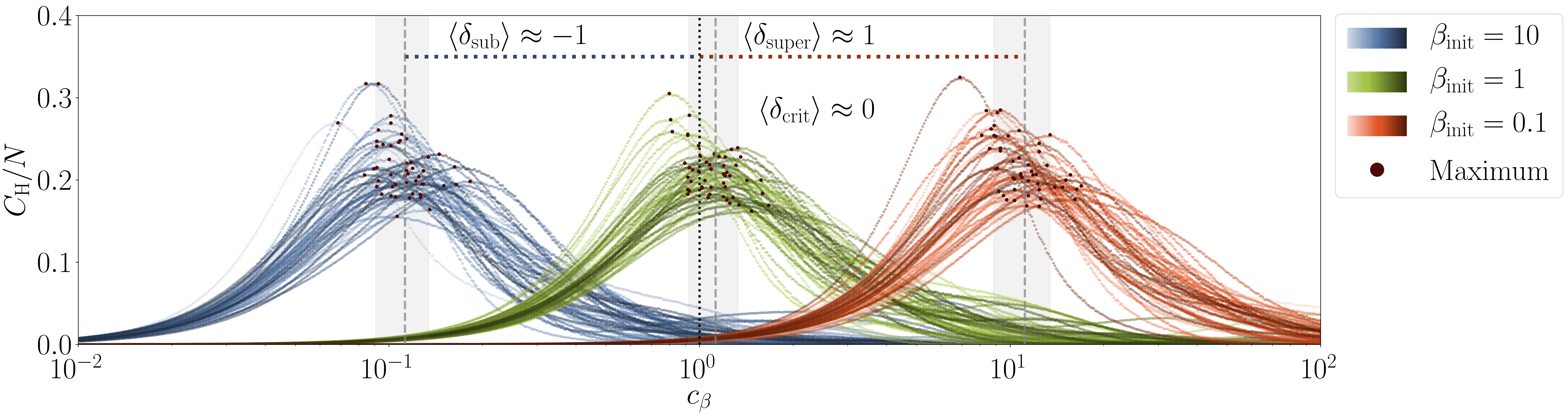}\vspace{-.8em}
}
\caption{The dynamical regime of a network can be calculated by finding a scaling factor $c_\beta$ of the inverse temperature that maximizes the heat capacity.
Heat capacity (Eq.~\ref{eq:Heat_Cap}) of the Ising  networks (Figure~\ref{fig:methods}) for 50 initially subcritical ($\beta_{\mathrm{init}} = 10$, blue), critical ($\beta_{\mathrm{init}} = 1$, green), and supercritical ($\beta_{\mathrm{init}} = 0.1$, red) organisms as a function of $c_\beta$.
For each organism it reaches the maximum (marked by a dot) at individual values $c_\beta = c_\beta^\mathrm{crit}$.
Dynamical regime $\delta = \log (c_\beta^\mathrm{crit})\approx - \log (\beta_{\mathrm{init}}).$ The displayed populations are unevolved and the resulting dynamical regimes closely correspond to their respective $\beta_{\mathrm{init}}$.}
\label{fig:heat_cap}
\end{figure}

We use the approximation of the
heat capacity from statistical physics to derive a measure of an organism's dynamical regime (sub-, super-, critical). Throughout the paper, we define the state of the organisms relative to the order/disorder transition.
In our finite system, we estimate the putative divergence point by changing the inverse temperature $\beta$ multiplying it with a scaling constant $c_\beta$.
This change of temperature influences how likely the state of the neurons will flip (Eq.~\ref{eq:spin_flip}), and thus change the equilibrium distribution of energies $e$ (Eq.~\ref{eq:ising_net_energy}). We search for a $c_\beta$ that maximizes value of the heat capacity $C_H(c_{\beta})$, defined as:
\begin{eqnarray}
    \label{eq:Heat_Cap}
    C_H(c_{\beta}) = \frac{1}{T^2} \textrm{Var}(e) = c_{\beta}^2  \beta^2  \textrm{Var}(e).
\end{eqnarray}
We define the $c_\beta^{\mathrm{crit}} = \underset{c_{\beta}}{\mathrm{argmax}}\, C_H(c_{\beta}) $. An analogous procedure was used in \citep{tkacik2015thermodynamics}. We then define the distance of the network from the critical point by the logarithm of the scaling factor required to bring the network to criticality, $\delta = \log(c_\beta^{\mathrm{crit}})$.
In our case, due to the asymmetric connectivity matrix and non-equilibrium nature of the system, the procedure should be seen as an approximation of the actual heat capacity, resulting in a proxy for critical point. More details on it in the Results section~\ref{Sec:Crit}.
For unevolved organisms whose connectivity matrices are initialized from the uniform distribution $\mathcal{U}(-1, 1)$ (first generation, Figure \ref{fig:heat_cap}), the relationship between $\beta_{\mathrm{init}}$ and $\delta$ can be approximated by:

\begin{eqnarray}
\delta \approx \log \frac{1}{\beta_{\mathrm{init}}} = - \log \beta_{\mathrm{init}}, \: \beta_{\mathrm{init}} \in [0.1, 10]
\label{eq:beta_delta_approx}
\end{eqnarray}

On a technical note, models in subcritical states can take increasingly large amounts of time to escape from a local optimum into a global optima, which can result in numerical divergences in our estimates of the specific heat $C_H(c_{\beta})$. To avoid this issue we employ an annealing method when calculating the specific heat at a given temperature, by first starting at a much higher temperature and then gradually lowering it down to the target temperature. During this process the sensor neurons are kept fixed according to values that the agents had observed and which are saved during training. This method has the benefit of being scalable especially for larger networks which may have very frustrated connections that take exponentially longer to equilibriate. 

During evolution, the distance from the critical point can (and will) change from its initialized state, and we must calculate its specific heat $C(\beta)$ as a function of the temperature scaling parameter $c_\beta$ in order to find its maximum and obtain an estimated distance to criticality.

\section{Results}
We perform extensive numerical experiments to investigate the properties of the dynamics of evolving Ising network agents and present the relationships between dynamical states, criticality and evolutionary fitness. 
However, before we present results on these dynamical states, we validate that our measures of criticality behave as expected by considering a generalized Ising model that is conceptually in between the classical Ising model and our Ising network agents. 

\subsection{Criticality in the generalized Ising model}\label{Sec:Crit}
\begin{figure}\centering
\makebox[\linewidth]{
\begin{tabular}{ c c }
\textbf{A} & \textbf{B} \\

 \includegraphics[height=5cm]{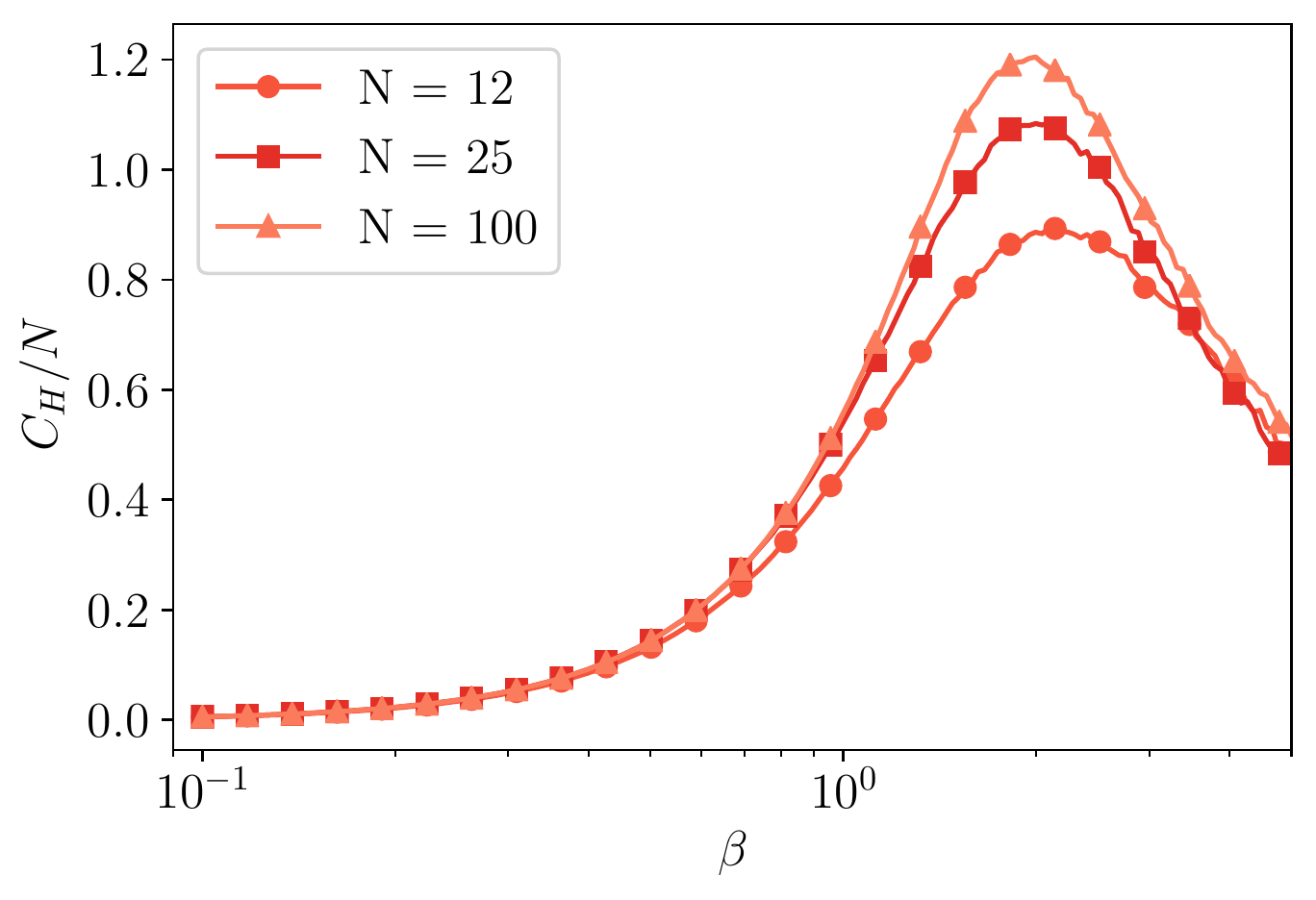} & \includegraphics[height=5cm]{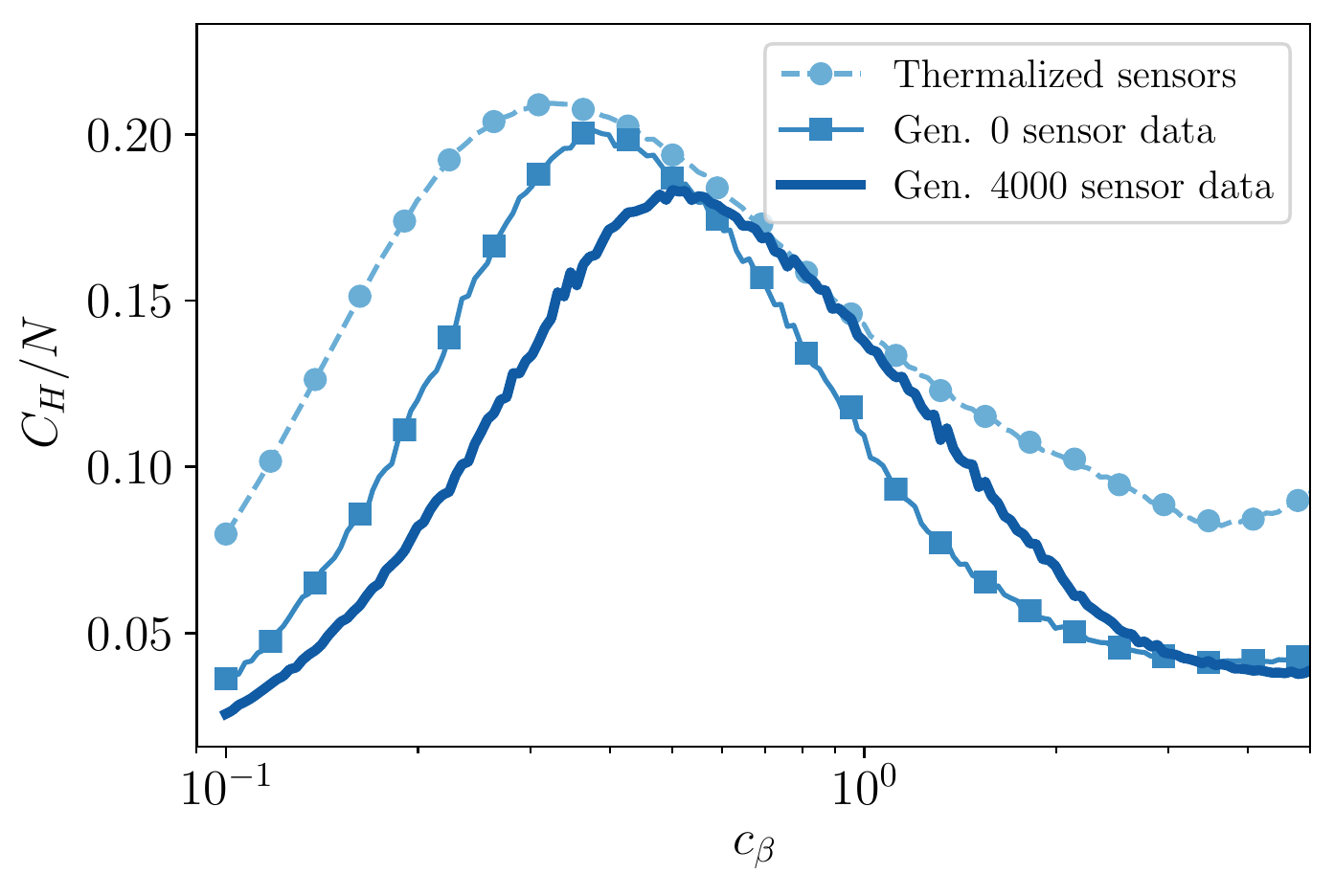}
\end{tabular}\vspace{-1em}}
\caption{Numerical indication of the presence of the critical point in our non-equilibrium generalized Ising model. \textbf{A:} A scaling analysis is done using ensembles of random networks with a similar architecture to the evolving agents. The peaks of the specific heat grows with the size of the system and its location converges towards a $\beta \approx 1.5$, a property commonly expressed in critical systems. \textbf{B:} 54 independent, fit agents, evolved for 4000 generations. The specific heat dependence on the inverse pseudo-temperature with thermalized sensor neurons (equilibrium model, light color), with the sensor data clipped and drawn from the distribution gathered at the 0th generation (darker shade), and  with sensor data from the 4000th generation (darkest line). 
The peak of the specific heat shifts slightly depending on the method.
}
\label{fig:criticality_analysis} 
\end{figure}

While the classical 2D Ising model and its critical point and universality class are well studied, extrapolating these results to different models and particularly to non-equilibrium systems is generally not possible and has to be checked individually for every variation of the model.
Particularly, in this paper, the controller of the agents is a neural network that represents a generalized Ising model with all-to-all connectivity, as opposed to a regular lattice, and with both positive and negative real-valued weights. Furthermore, the controller neural network receives sensory inputs that perturb the model away from equilibrium, while being at equilibrium is the central requirement for the derivation of the critical points in the 2D Ising model. Finally, controller networks evolve very specific connectivities via the selection pressures of the world/task they are embedded in, this precludes a scaling analysis on the specific evolved networks to measure if their thermodynamic properties exhibit scale-free, and therefore critical, behaviours. As a compromise, we instead do a scaling analysis on random networks with a similar architecture as our controller networks. We generate ensembles of random networks of sizes $N = 12, 25, 100$, each having 1/3rd of its neurons designated as sensor neurons and another 1/3rd as motor neurons, prohibiting connections between motor and sensor groups.
We normalize the weights by the Frobenius norm of the connectivity matrix. We then calculate the heat capacities of these models, where the sensor neurons are given values from the uniform distribution on $[-1, 1]$, shown in Figure~\ref{fig:criticality_analysis}A. It can be seen that the specific heat of these models tend to peak  around $\beta \approx 1.5$, showing that our normalization captures the changes of the peak location with the system size.  The maximum value of the specific heat is growing with system size, implying the existence of a critical point.

The next complication in our methodology is the fact that the sensor neurons perturb the system away from equilibrium at each time step, and therefore make our analysis of criticality more difficult. To better understand the implications of this feature in our model, we can compare how the specific heat of a network changes depending on the statistics of the sensory input and the possibility to thermalize its sensor neurons. We consider the most fit agent from 54 independent simulations of 4000 generations evolved with the GA. We calculate their specific heats for various inverse temperatures three different ways and average across the agents. In the first method, we thermalize the sensor neurons and treat them identically to the rest of the neurons in the model. This results in the equilibrium model that is different from the Ising model only in the features of its connectivity matrix, see Figure \ref{fig:criticality_analysis}B lightest curve. We repeat the same calculations with clipped sensor neurons drawn from the distribution of sensor data gathered in its final generation (4000) which we save throughout the lifetime of the agents. This is the `effective' specific heat of the embodied model as it interacts with its environment, and it is how we defined the state of the agents in the rest of the paper, Figure \ref{fig:criticality_analysis}B darkest curve. Finally, we calculate the specific heat using the sensor data from the 0th generation, where due to the unevolved state of the agents, the sensor data is less diverse. The difference between the thermalized specific heat and the `effective' specific heat is a slight shift in the location of the peaks. Furthermore, it can be seen that the evolved agents are closer to their maximal susceptibility when their specific heat is calculated from the environment they are actually embedded in. In other words, if we were to calculate the specific heat of these agents using the equilibrium model by discarding sensor data, we would systematically over-estimate how subcritical a network is due to the fact that the environment is interacting with the agent and vice-versa. 

\begin{figure}\centering
\makebox[\linewidth]{
\begin{tabular}{ c c c }
\textbf{A.} Simple Task & \textbf{B.} Simple Task & \textbf{C.} Hard Task\\
$12$ Neurons & $28$ Neurons & $12$ Neurons \\
 \includegraphics[height=3.5cm]{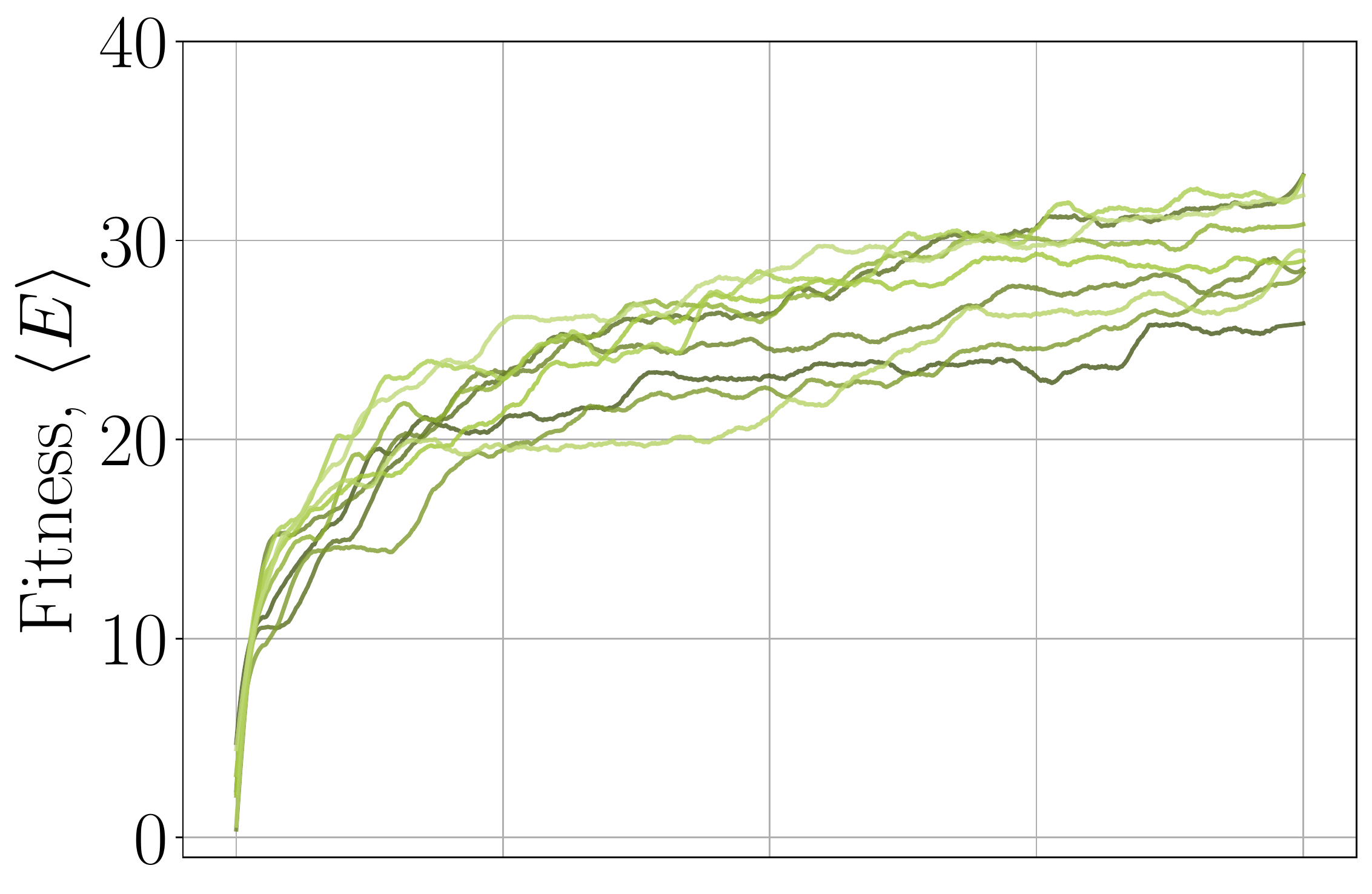} & \includegraphics[height=3.5cm]{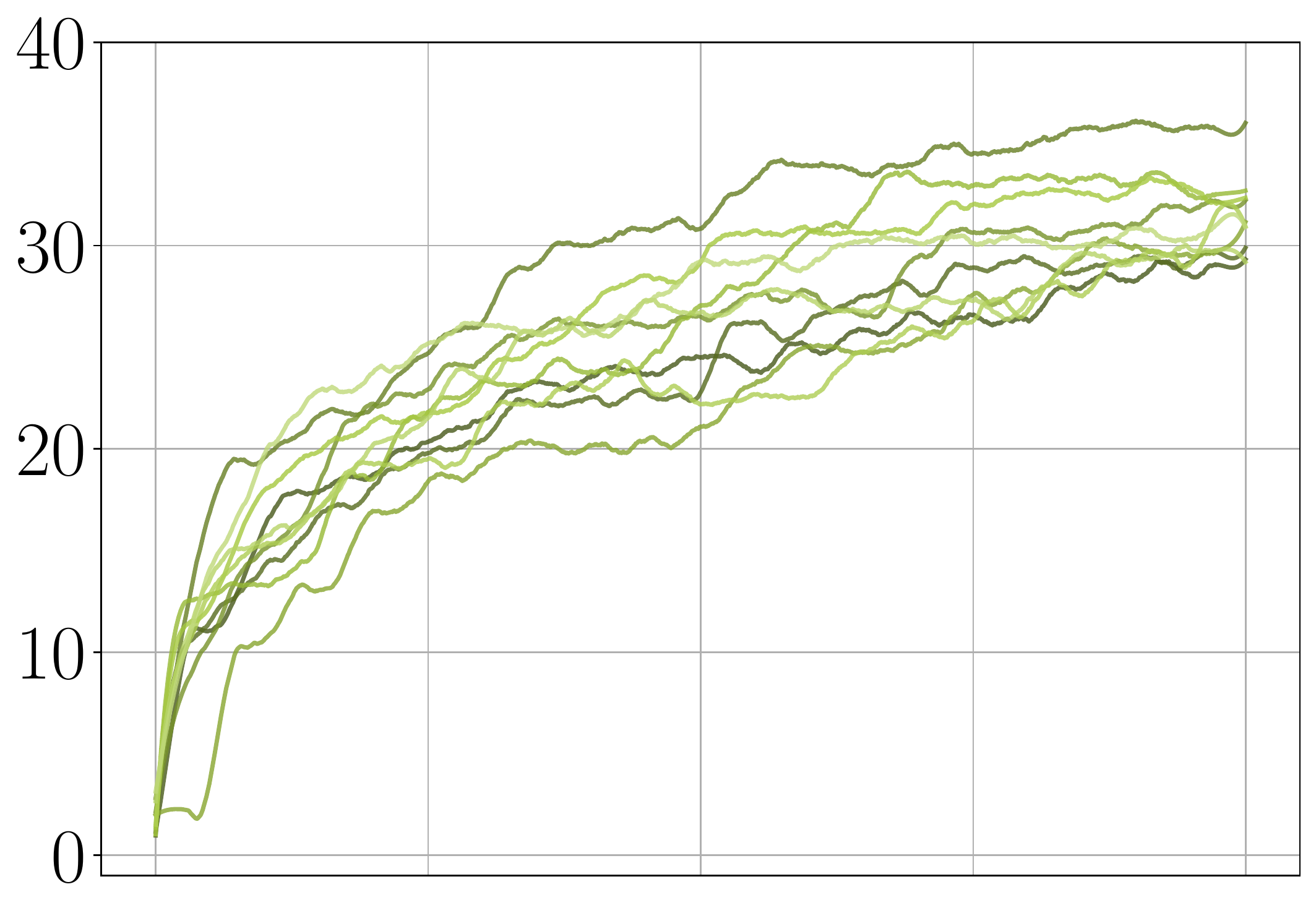} & \includegraphics[height=3.5cm]{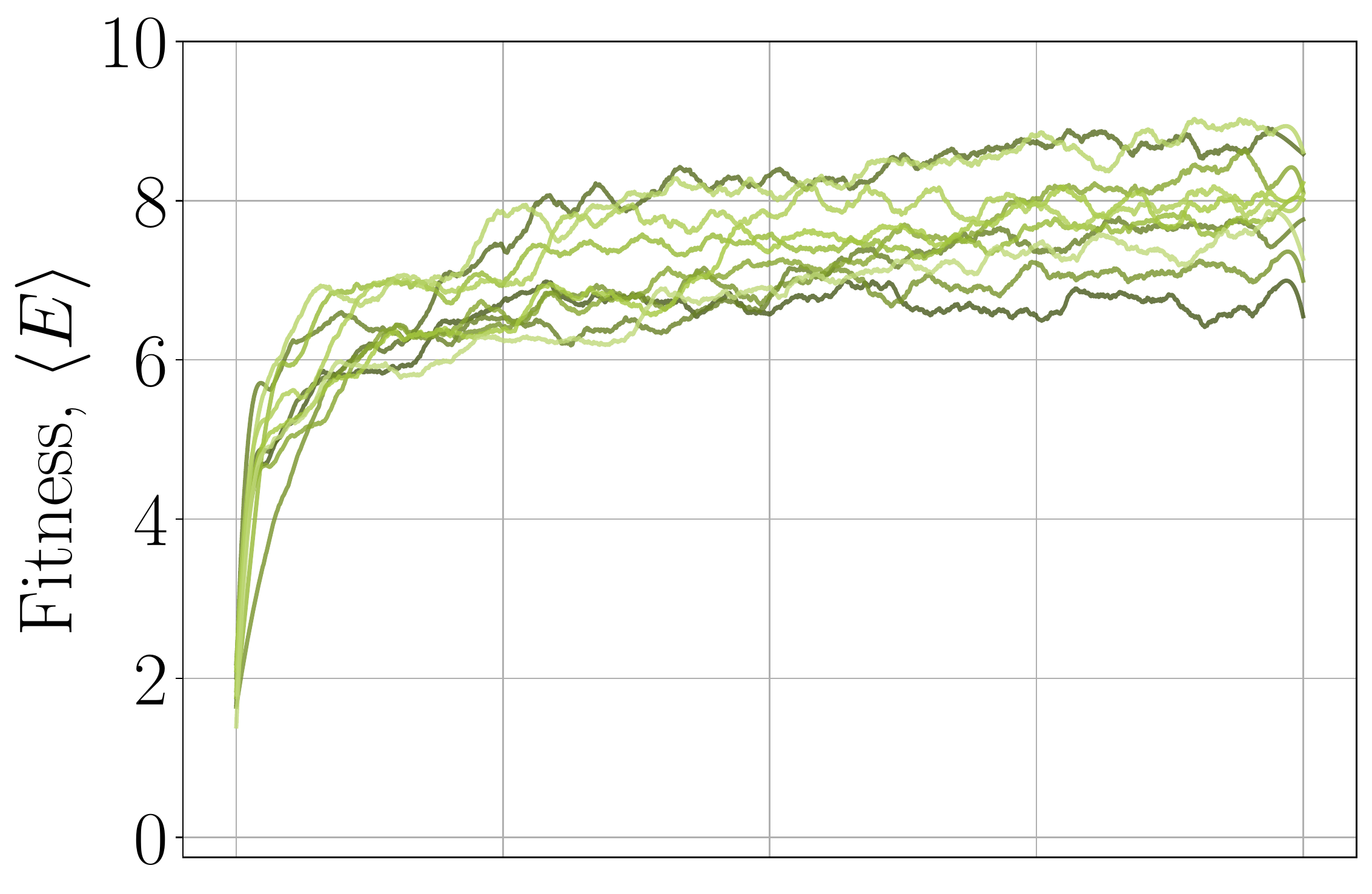} \\
 \includegraphics[height=4.0338cm]{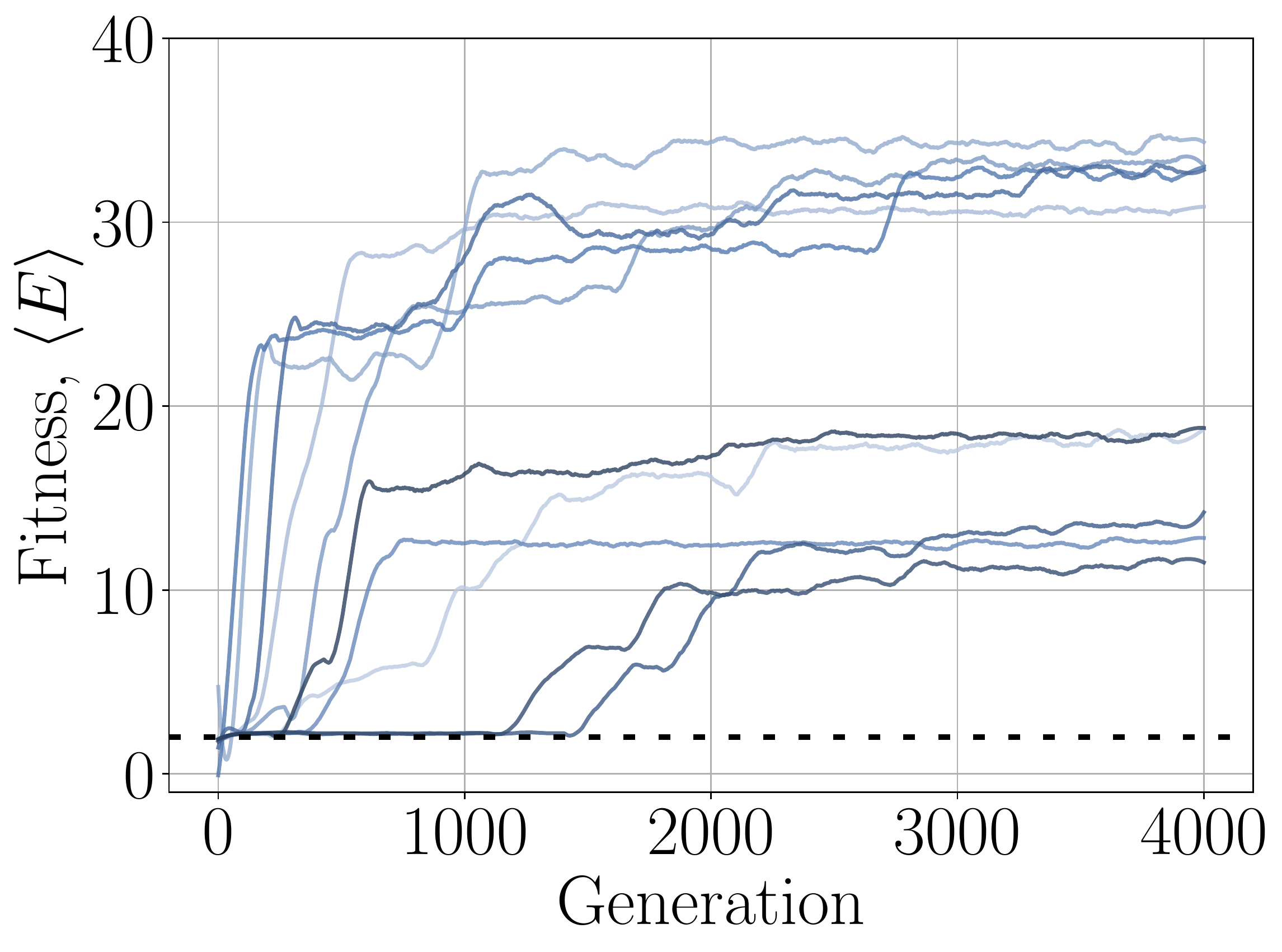} & \includegraphics[height=4.0338cm]{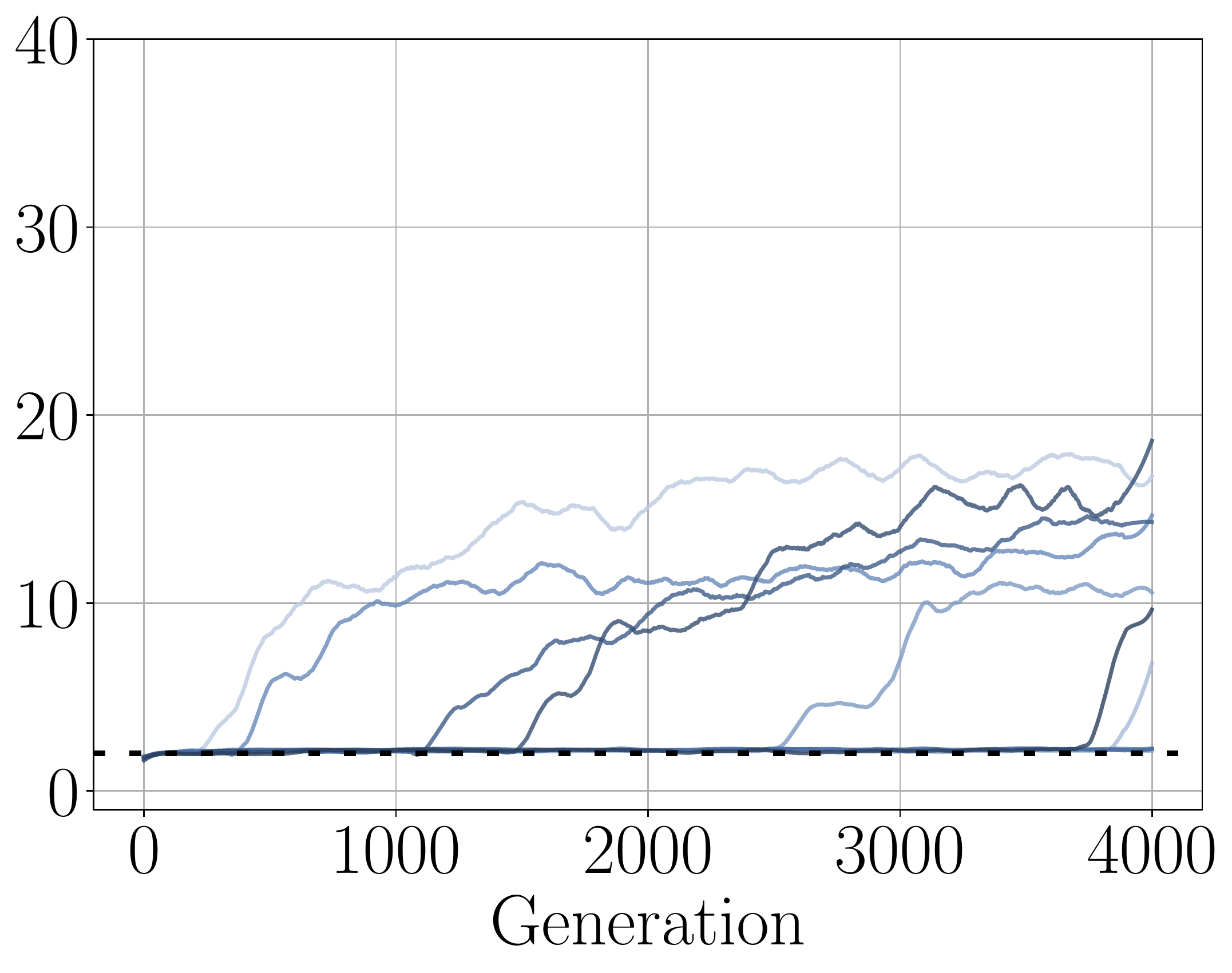} & \includegraphics[height=4.0338cm]{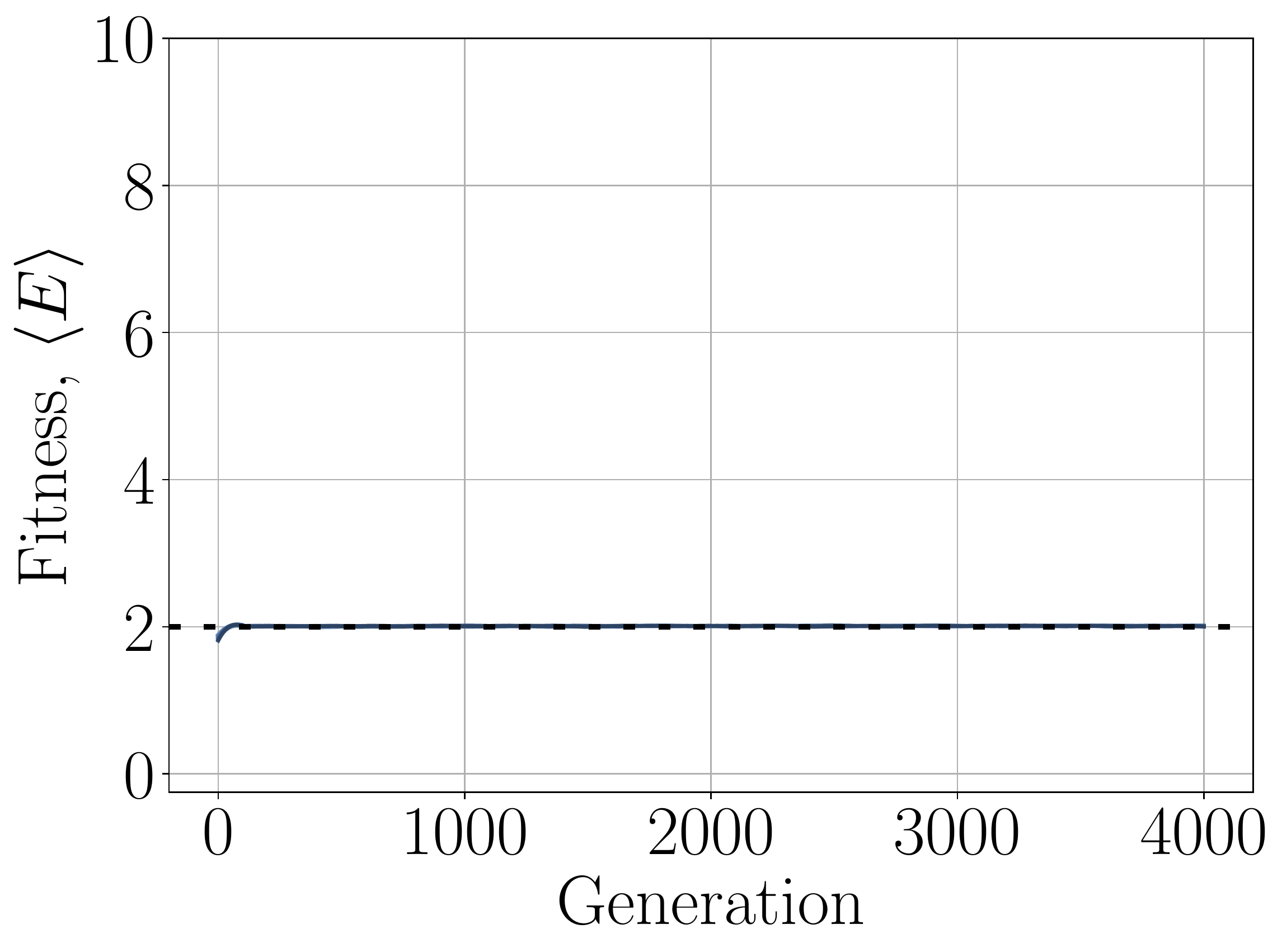}
\end{tabular}\vspace{-1em}}
\caption{Critically initialized populations can be successfully evolved in different circumstances, whereas for subcritically initialized populations, a harder task or an increased system size can significantly disrupt evolutionary dynamics.
For each panel 10 initially critical ($\beta = 1$, green, top row) or initially subcritical ($\beta = 0.1$, blue, bottom row) populations evolve for 4000 generations.
The dashed line at fitness = 2 in the subcritical panels correspond to the organisms' initial energy. It can be seen in the hard task with 12 neuron (bottom right) that the network is unable to achieve a fitness above its starting value.}
\label{fig:fitness_plots_vs_generation} 
\end{figure}

\subsection{Convergence of evolution}
Populations of different initial states follow distinct evolutionary strategies and most are able to solve the standard foraging task when evolved with the genetic algorithm (GA). Within the range of $\delta_{\mathrm{init}} \in [-1,1]$ as in the previous paper~\citep{Prosi}, all populations converge to a good fitness, but for $\delta_{\mathrm{init}} \ll -1$ or $\delta_{\mathrm{init}} \gg 1$ the GA sometimes cannot find suitable solutions. 
We observe evolution for 4000 generations for populations initiated between the ranges of subcritical ($\beta_{\mathrm{init}} \approx 32$, $\delta \approx -1.5$), critical  ($\beta_{\mathrm{init}} = 1$, $\delta \approx 0$) and supercritical ($\beta_{\mathrm{init}} \approx 0.03$, $\delta \approx 1.5$) regimes.
Critical populations begin to rapidly gain fitness from the first generation in every independent simulation run. (Figure~\ref{fig:fitness_plots_vs_generation}A).
The gradual and stable increase of fitness of the initially critical population suggests that successful hill climbing on the fitness landscape is taking place.
In contrast, for subcritical populations, fitness mainly evolves via random jumps and only about half of the simulations reach the same fitness as the critical populations after 4000 generations (Figure~\ref{fig:fitness_plots_vs_generation}A, video: \url{https://vimeo.com/547613948}).
Such fitness dynamics indicate a random search strategy which often leads to a population getting trapped in a local maxima for extended periods of time.
Confirming the previous observations by~\citet{khajehabdollahi2020sinas_paper}, we see that supercritical populations, after an initial random period follow the same path as the critical ones, though highly supercritical models with $\delta > 1$ can sometimes struggle to find solutions.

\begin{figure*}\centering
\figuretitle{Genetic Algorithm (GA)}
\makebox[\linewidth]{
\begin{tabular}{ c c }
\textbf{A.} Simple Task & \textbf{B.} Hard Task\\
 \includegraphics[height=1.8in]{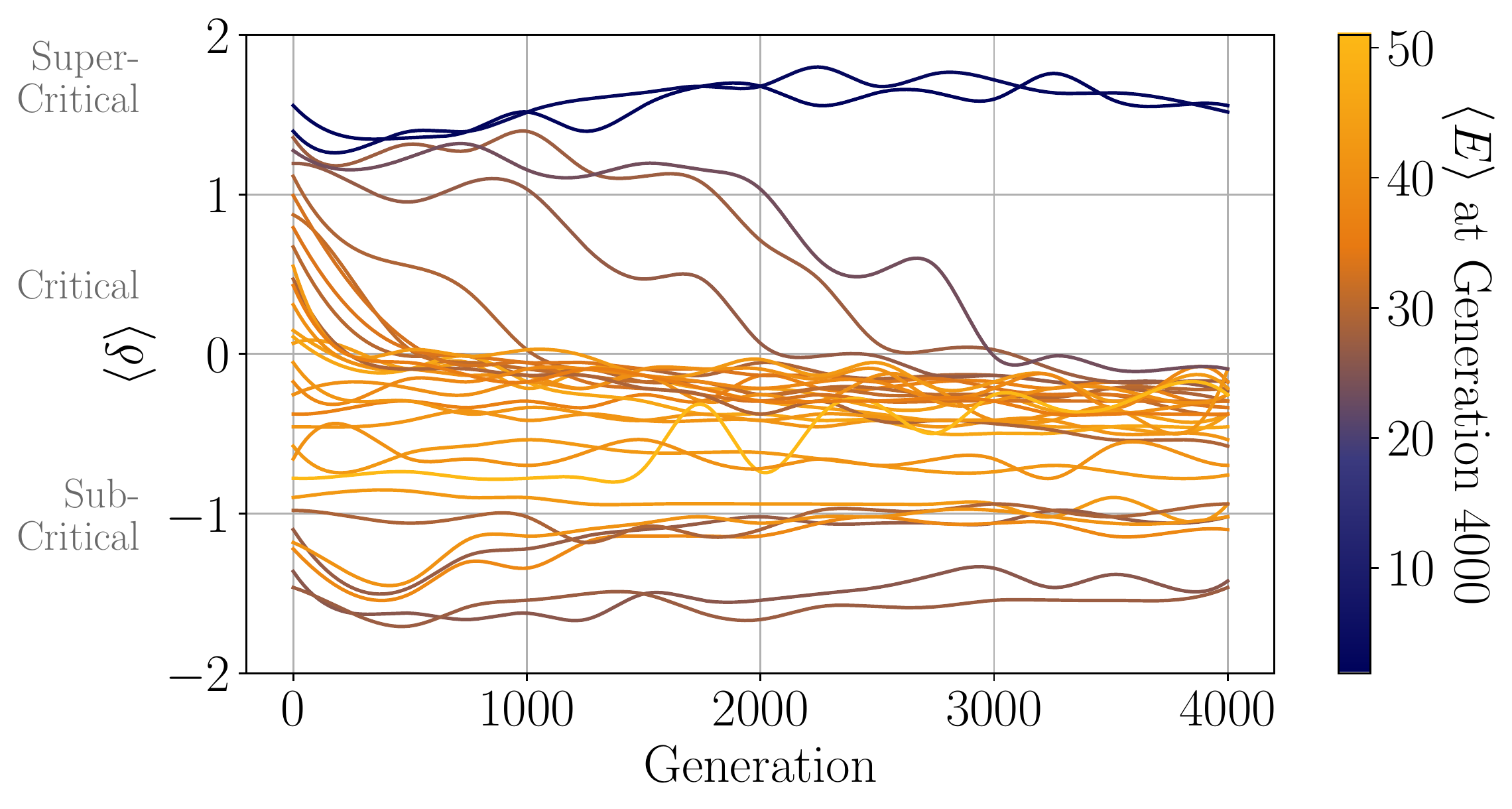} & \includegraphics[height=1.8in]{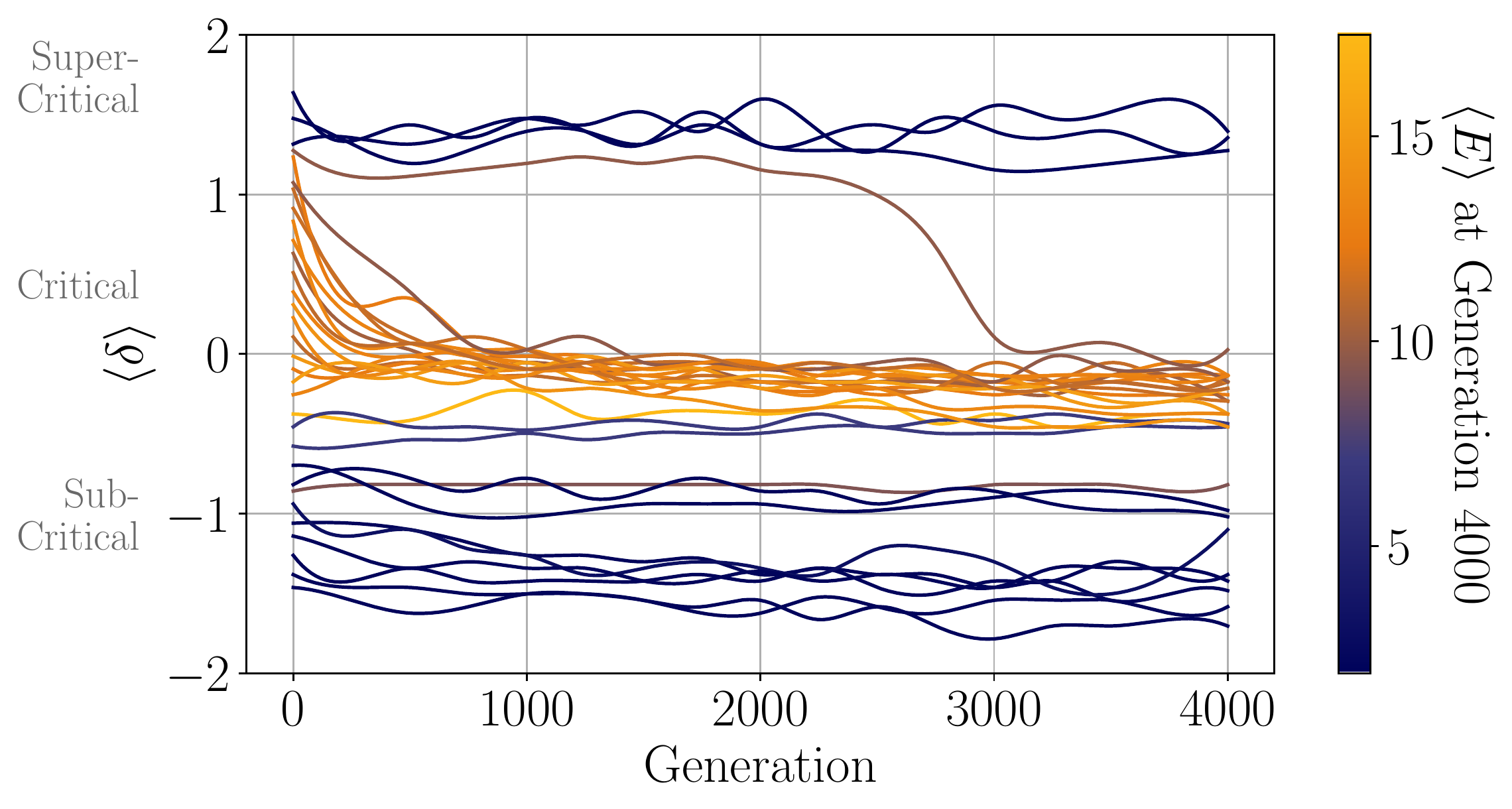}
\end{tabular}}\vspace{-1em}
\caption{Changes in the distance to criticality over the course of evolution.
32 populations initiated at various distances to criticality ($\delta$ between -1.5 and 1.5) and evolved on a \textbf{A:} simple task and \textbf{B:} hard task.
The color indicates their fitness at generation 4000. Most populations with $\delta < 0$ remain at fitness 2 for the hard task, as well as a few simulations with $\delta > 1$ signifying no evolutionary progress.
}
\label{fig:delta_vs_generation_GA}
\vspace{1.5em}

\makebox[\linewidth]{
\begin{tabular}{ c c }
 \includegraphics[height=1.8in]{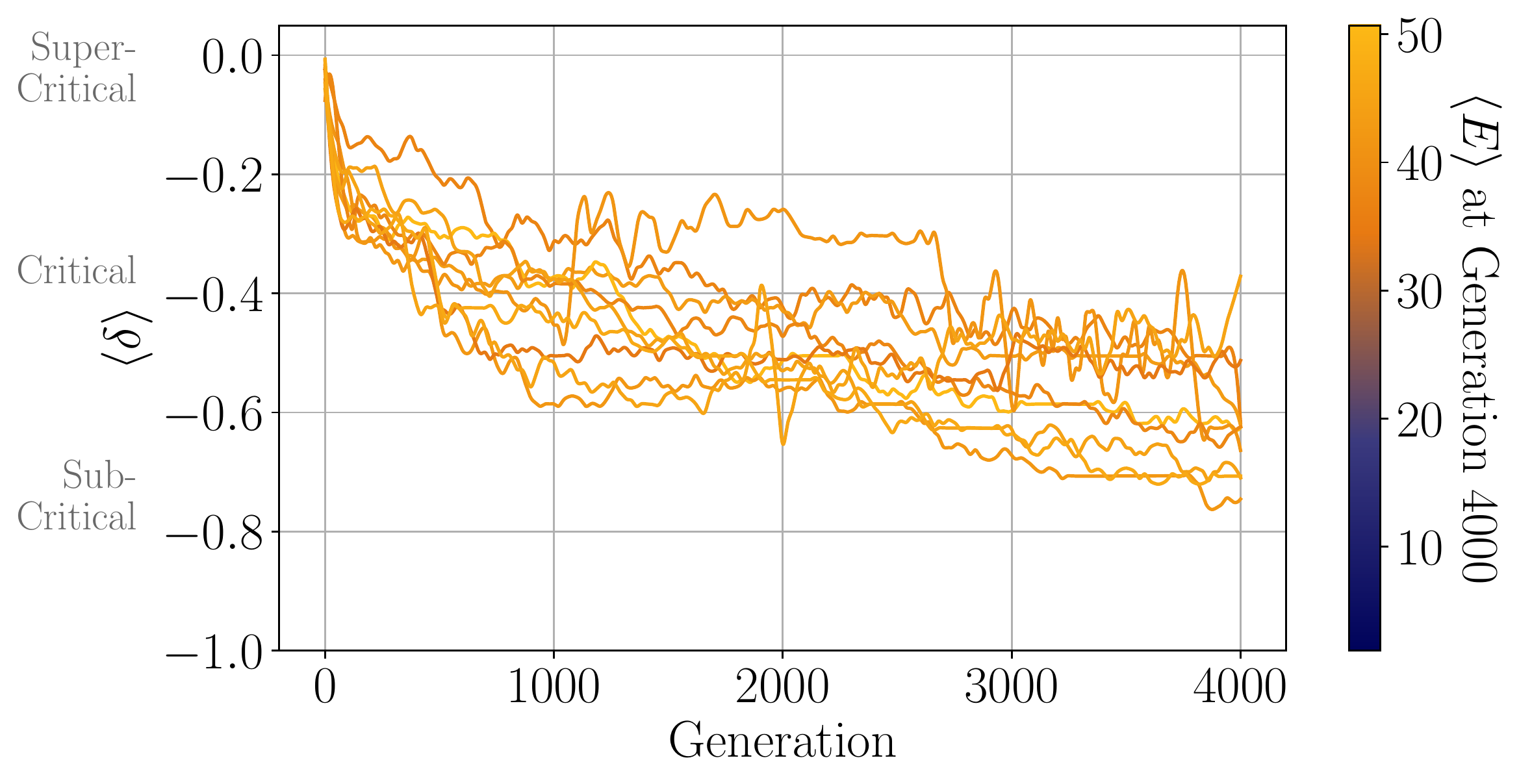} & \includegraphics[height=1.8in]{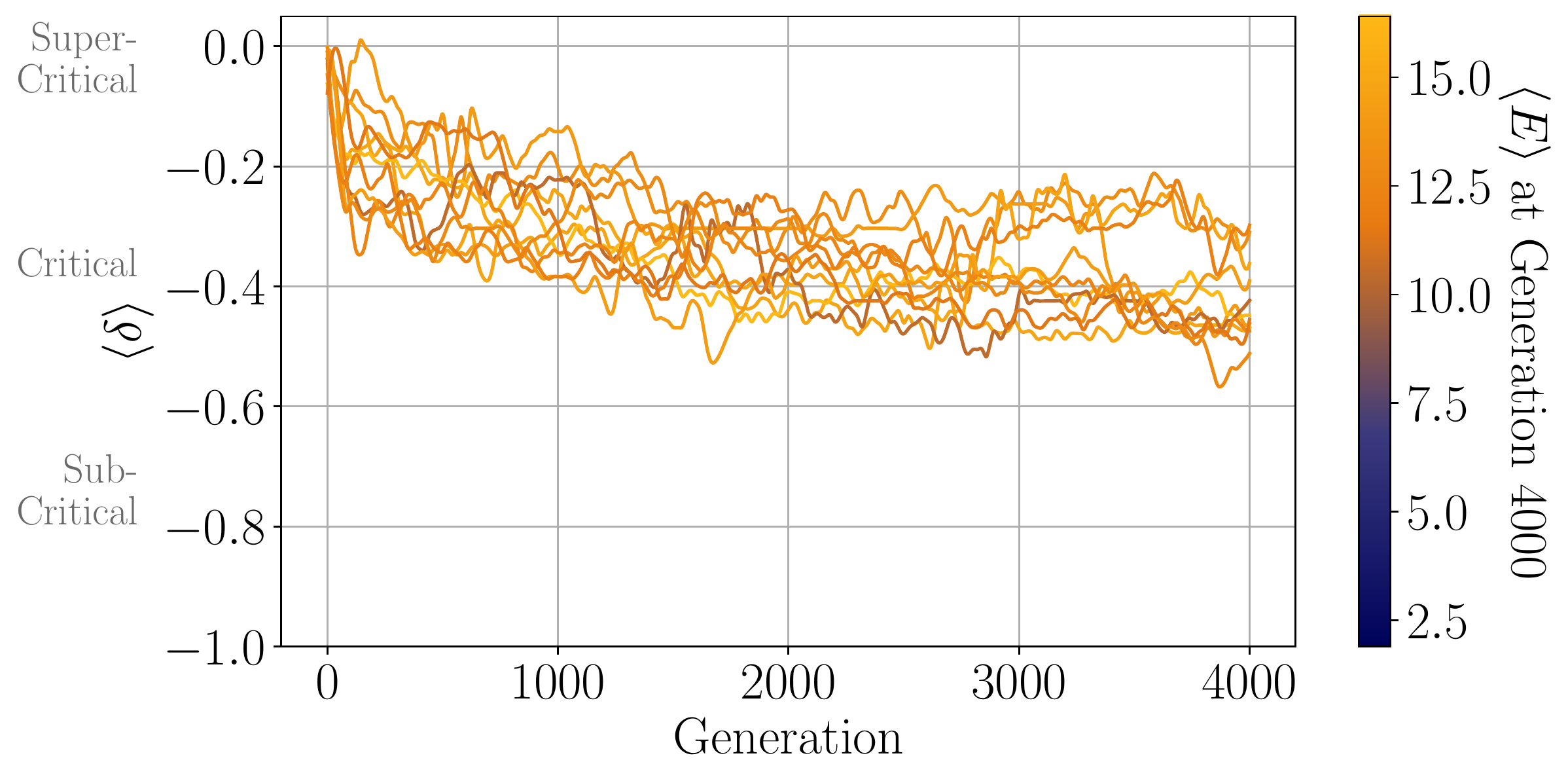}
\end{tabular}}\vspace{-1em}
\caption{State dynamics of 10 populations initialized at $\delta = 0$ and evolved on a \textbf{A:} simple task and \textbf{B:} hard task. All population solve the corresponding task but the populations trained on the hard task evolve to have $\delta$ values closer to criticality.
}
\label{fig:delta_vs_generation_b0_GA}
\end{figure*}


\begin{figure*}\centering
\figuretitle{Evolution Strategy (ES)}
\makebox[\linewidth]{
\begin{tabular}{ c c }
\textbf{A.} Simple Task & \textbf{B.} Hard Task\\
 \includegraphics[height=1.8in]{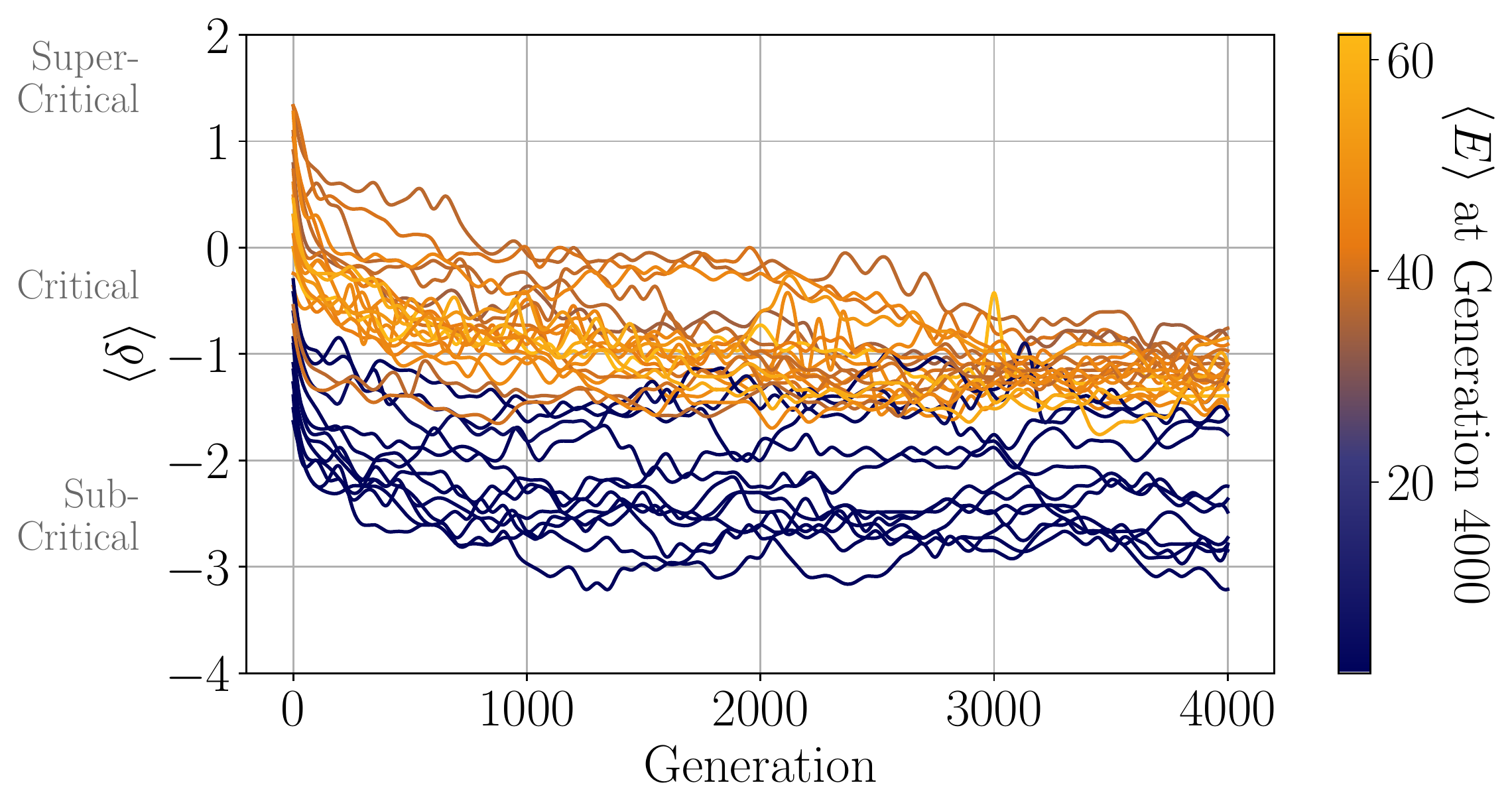} & \includegraphics[height=1.8in]{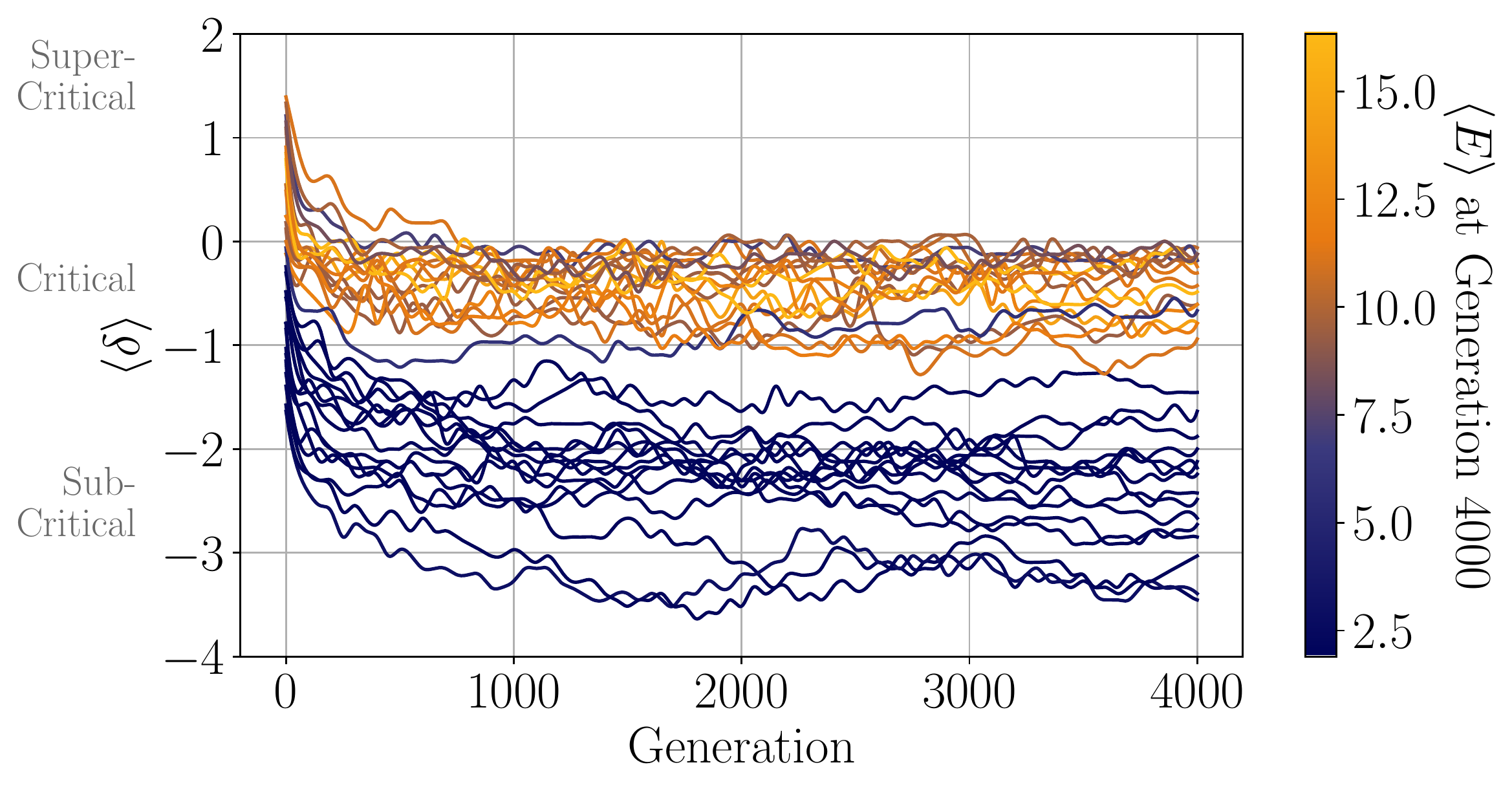}
\end{tabular}}
\vspace{-1em}
\caption{Changes in the distance to criticality over the course of evolution .
32 populations initiated at various distances to criticality ($\delta$ between -1.5 and 1.5) and evolved on a \textbf{A:} simple task and \textbf{B:} hard task.
The colour indicates their fitness at generation 4000. Similar, but more dramatic, than the results from the GA, populations initialized with $\delta < 0$ suffer greatly in their ability to discover optimal solutions in both tasks. The ES can also be observed to have an overall tendency to become subcritical even when lacking selection pressure.
}
\label{fig:delta_vs_generation_NES}
\vspace{1.5em}

\makebox[\linewidth]{
\begin{tabular}{ c c }
 \includegraphics[height=1.8in]{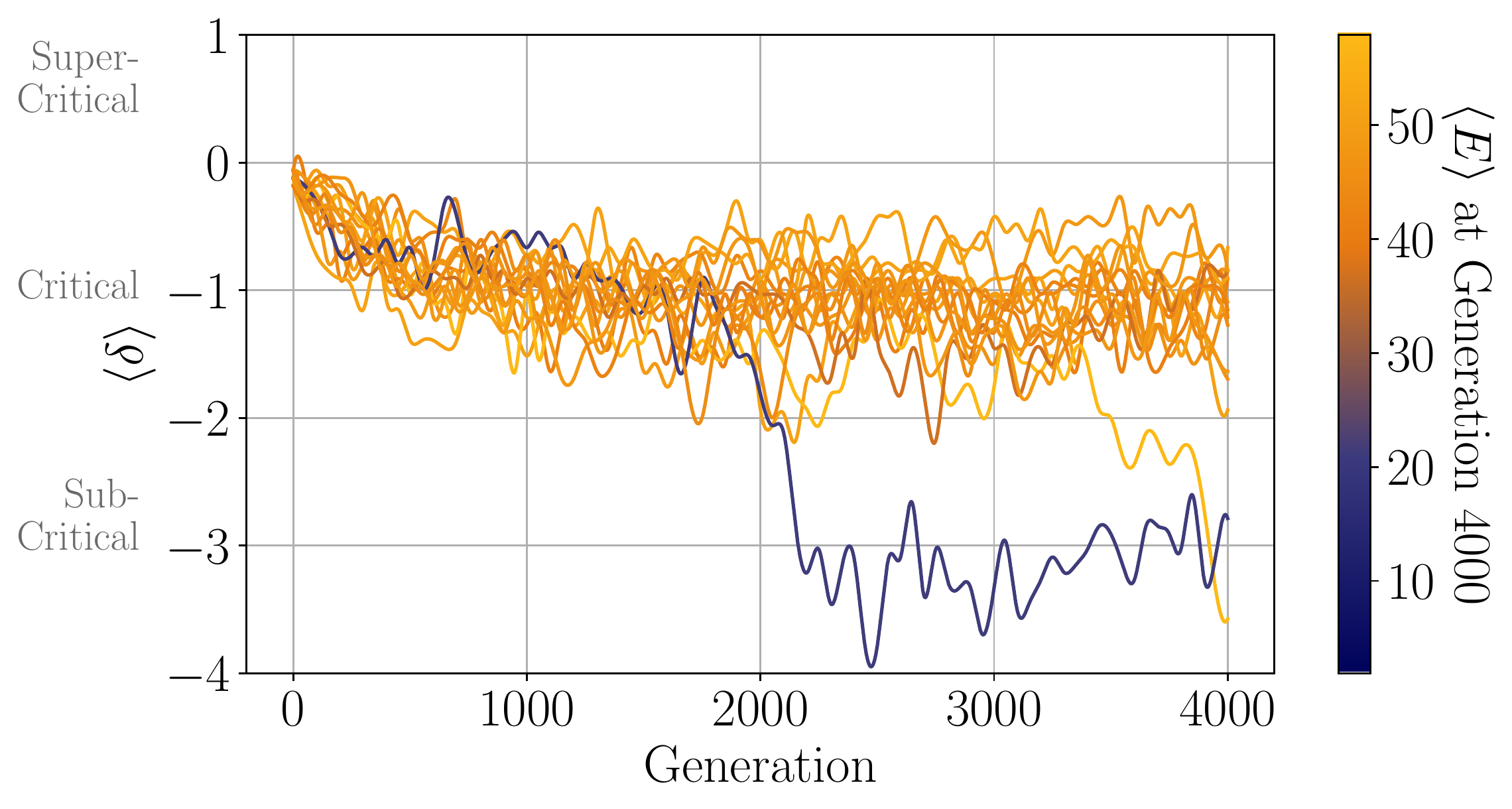} & \includegraphics[height=1.8in]{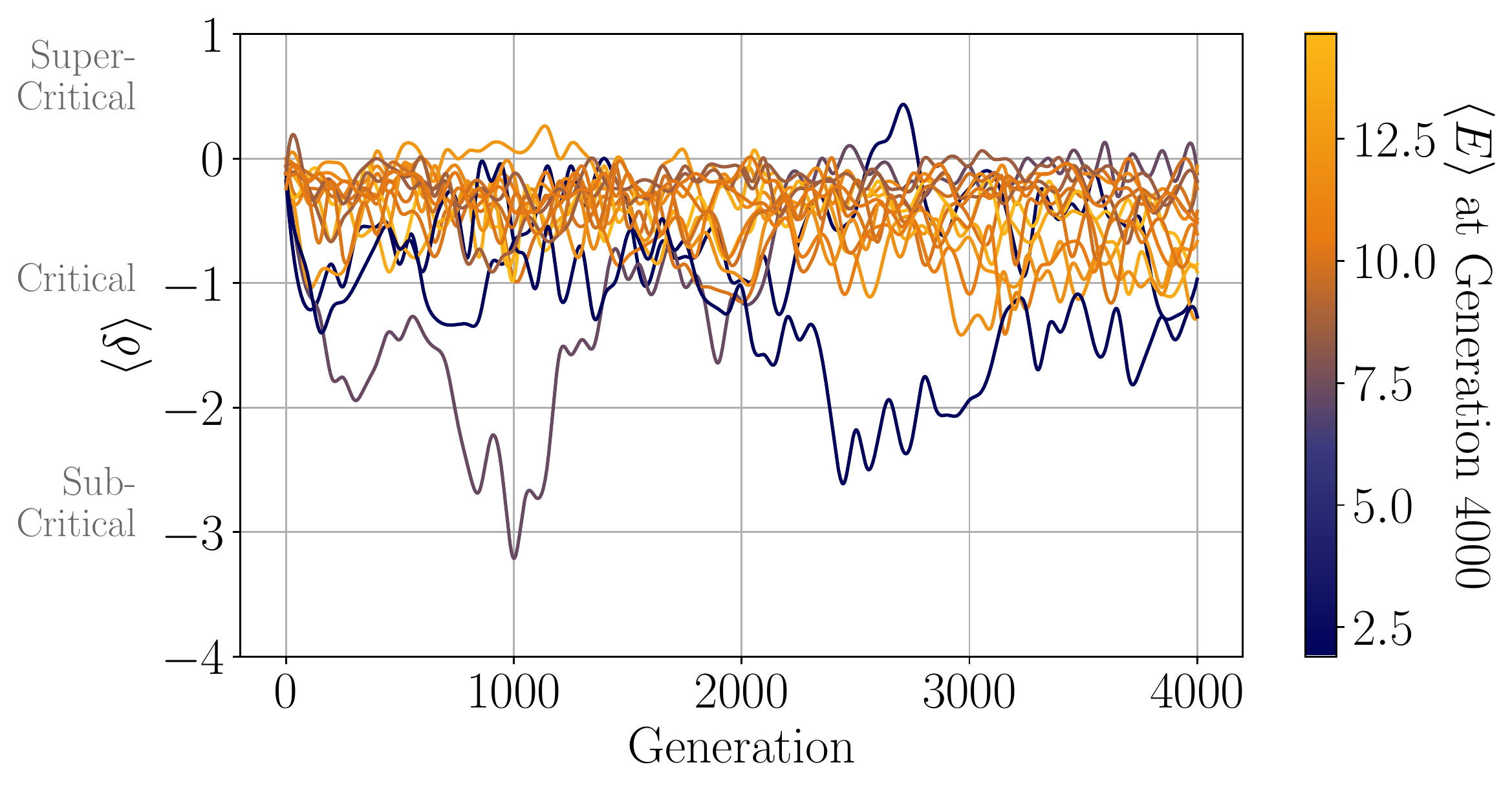}
\end{tabular}}\vspace{-1em}
\caption{16 populations initialized at $\delta = 0$ and evolved on a \textbf{A:} simple task and \textbf{B:} hard task. The populations trained on the hard task evolve to have $\delta$ values closer to criticality, similar to the results from the GA, albeit at larger values of $\delta$.
}
\label{fig:delta_vs_generation_b0_NES}
\end{figure*}

\begin{figure}\centering
\makebox[\linewidth]{
\begin{tabular}{ c c }
\textbf{A.} GA & \textbf{B.} ES\\
\includegraphics[width= 0.5\textwidth]{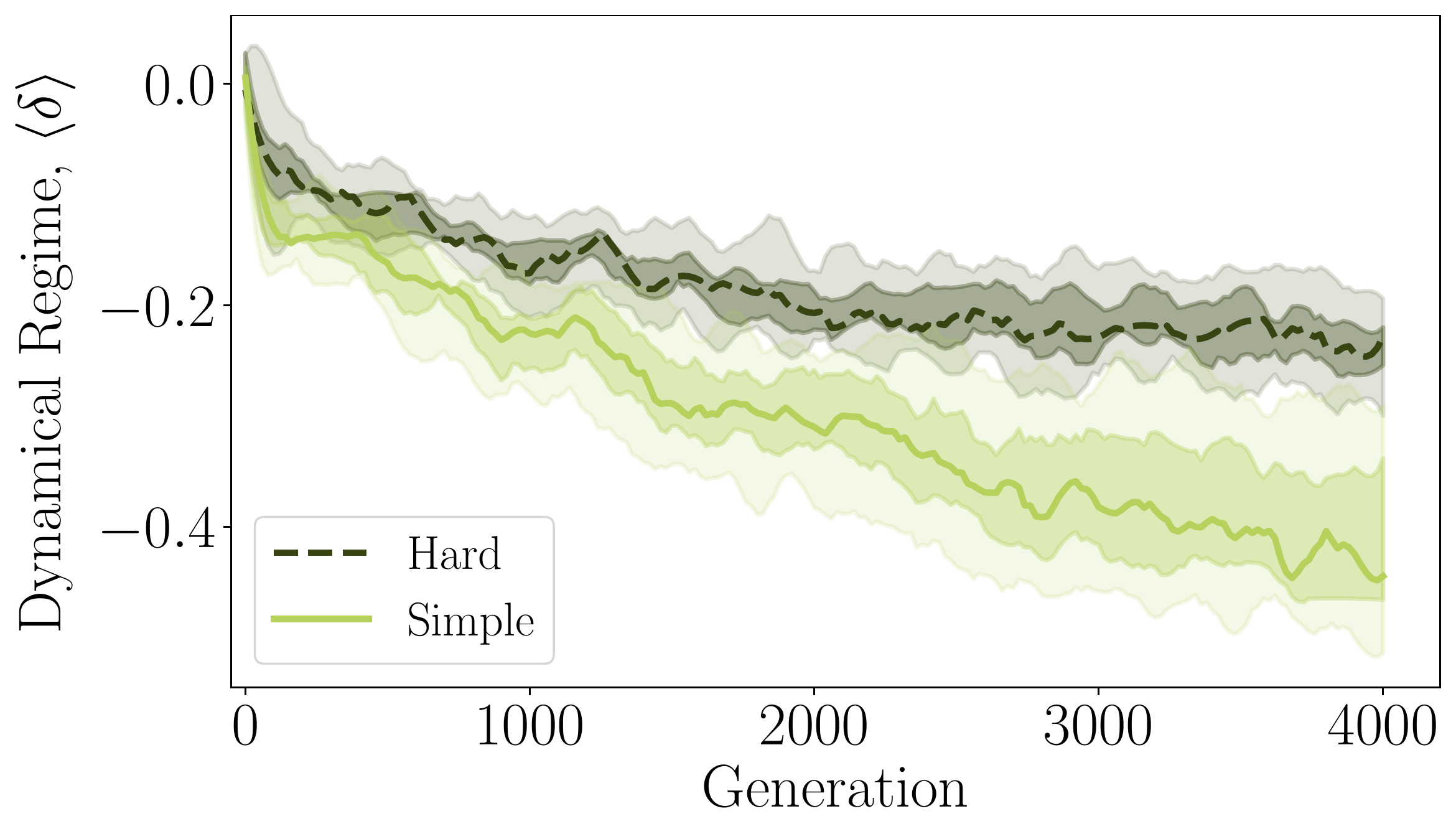} \hfill &
\includegraphics[width= 0.5\textwidth]{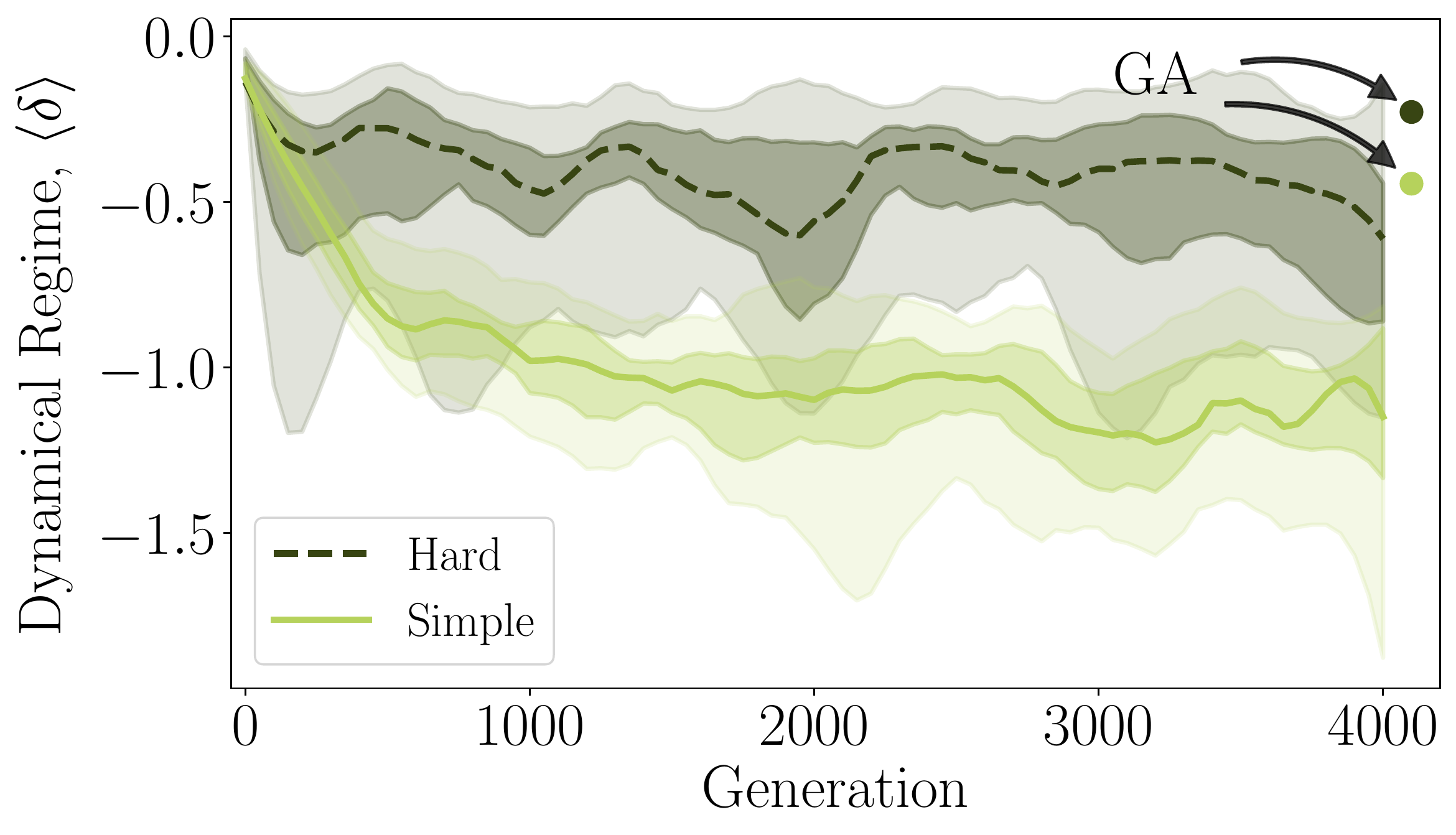}
\end{tabular}}\vspace{-1em}
\caption{The dynamical regime of the initially critical population remains closer to the critical regime ($\delta=0$) in the harder task than in the simple task throughout evolution, for both algorithms (tested for the final generation 4000 with Mann-Whitney U test,  $p = 2.2 \times 10^{-6}$ and $p = 4.4 \times 10^{-3}$, for the GA and ES, respectively). 10 and 16 populations for the GA~(\textbf{A}) and ES~(\textbf{B}), respectively, are initiated in the critical regime end evolve for 4000 generations. The 15\textsuperscript{th} and 85\textsuperscript{th} percentiles are lightly shaded, and the 33\textsuperscript{rd} and 67\textsuperscript{th} percentiles are more heavily shaded. 
}
\label{fig:simple_vs_hard_task}
\end{figure}

For successfully evolvable populations, moderate changes in the complexity of the control network should not destroy the ability of the EA to reach a good fitness.
We test the differences in evolvability for initially critical and subcritical populations by changing the size of the network from 12 to 28 neurons.
As in smaller networks, the initially critical populations rapidly evolve for all initial conditions.
By generation 4000 they even reach a slightly higher fitness than populations controlled by smaller networks and evolved for the same number of generations (Figure~\ref{fig:fitness_plots_vs_generation}B).
However, initially subcritical populations do not reach even half of their original fitness.
We observe the same difference between the dynamical regimes when we increase the task's complexity requiring organisms to slow down to almost zero velocity in order to consume food particles (Figure~\ref{fig:fitness_plots_vs_generation}C, video: \url{https://vimeo.com/547615705}).
In this harder task, the evolved populations' maximal fitness is expected to be lower than for the simple task.
For the initially critical populations, we still observe the same hill-climbing dynamics. However, the initially subcritical populations stay at an energy level of precisely two.
This signifies that they do not use the originally supplied energy for moving and remain static throughout all 4000 generations, trapped in a local optimum.

Overall, we see that although in simple tasks all populations can converge to approximately the same fitness, there exists a significant difference between the initially subcritical and initially critical/supercritical populations.
Specifically, the convergence of evolution for critical populations is stable (all populations follow very similar fitness growth) and behave similarly regardless of network size or task complexity.
For subcritical populations, the evolutionary dynamics resembles random search, which fails to find solutions in high-dimensional cases or for more complex tasks.

\subsection{Evolution of the dynamical regime}
\label{subsec:evo_delta}

\fs{Next, we investigate how the dynamical state of the populations changes during evolution.}
To do so, we select a wide range of initial dynamical regimes ($\delta \in [-1.5, 1.5]$) and examine how the dynamics of populations initialized in each of these regimes change throughout evolution via the GA. 
Regardless of their initial dynamics, almost all populations that manage to find solutions converge to the subcritical regime, albeit with different distances from the critical point (Figure \ref{fig:delta_vs_generation_GA}). The populations that did not follow this convergence pattern were also ones that never discovered solutions to the task. We also observe that strongly subcritical populations ($\delta < -1$) and strongly supercritical populations ($\delta > 1$) generally achieve lower fitness in the simple task and are unable to solve the hard task.

A basin spanning the near-critical regime, from moderately subcritical to moderately supercritical  populations,  rapidly change  their  dynamical  regime  and  by  generation  4000 reach an intermediately subcritical state, whose $\delta$ we refer to as $\delta^* \approx -0.41$. This is a relatively broad dynamical regime whose evolutionary dynamics have different characteristics than the dynamics of deeply subcritical networks.

Deeply subcritical populations with $\delta\ll\delta^* $ remain at their initial regimes for the GA, demonstrating a lack of evolutionary mobility and consequently are more likely to obtain lower fitnesses, whereas subcritical populations initialized at higher $0 \geq \delta \geq \delta^*$  can still approach $\delta^*$ which is correlated with the ability to solve the underlying task with high fitness. Similarly, deeply supercritical populations also struggled to change their dynamical regime, however the supercritical populations that were able to optimize all converged to $\delta^*$, much like the near-critical populations (Figure~\ref{fig:delta_vs_generation_GA}).

Task complexity determines the dynamical regime where evolution converges to. Specifically, when trying to solve the hard task, the agents converge to a smaller distance from criticality than when solving the original task.
We check the evolution of the dynamical regime in both simple and hard tasks (Figure~\ref{fig:delta_vs_generation_b0_GA}).
We utilize the observation that almost all populations with an initial regime $\delta > \delta^*$ converge to similar values. Thus, we consider only initially critical populations.
We obtain the distribution of dynamical states by considering 10 independent runs of evolution in both tasks after 4000 generations. We take the mean of the top 30 most fit agents in each simulation, and perform a Mann-Whitney U test to confirm that the $\delta$ values for the hard task are larger than the simple task ($p < 10^{-5}$), i.e. closer to the critical value.
Specifically, the harder task results seem to consistently maintain a smaller distance to the critical point throughout evolution (Figure~\ref{fig:simple_vs_hard_task}.A).

\fs{Therefore, initiating an agent close to the critical regime is important when task complexity is unknown.}
We observe that the dynamical regime never changes towards supercriticality, but the subcritical convergence point can be at different distances from criticality. Thus, only starting near the critical point guarantees that the optimal dynamical state can be reached by evolution.

To verify that our results are not contingent on the specific implementation of GA, we run the same experiment using a different optimization method -- an Evolution Strategy (ES), as described in Sec.~\ref{sec:ES}. 
We re-run the experiments using ES and obtain qualitatively similar results. In Figure \ref{fig:delta_vs_generation_NES}, we observe that simulations near or above the critical point are able to discover high-scoring solutions in both the simple and the hard task. Furthermore, we once again observe that when initialized below a certain point, populations are unlikely to discover a good solution using the ES (even more than we observed for GA).

We observe that solutions to the simpler task converges to a more subcritical regime than for the hard task.
However,  the ES results in larger deviations form the critical point in the converged populations than the GA. 
This can be potentially attributed to the faster convergence of the dynamical state allowed by the ES.

We verify that the difference in the distance to criticality between simple and hard tasks is significant under both evolutionary algorithms, as shown in Fig.~\ref{fig:simple_vs_hard_task}B.
For each of the tasks, 16 independent populations of initially critical agents are evolved and the distribution of their $\delta$ values is presented.
 We compare the final $\delta$ values after 4000 generations and confirm ($p < 10^{-2}$, Mann-Whitney U test) that the simpler task converges to a more subcritical regime than the hard task for the ES as well. These results indicate that our findings are independent of the evolutionary algorithm used to solve the task.

\subsection{Comparison of evolutionary algorithms}
To understand how these two different families of evolutionary algorithms function differently, we compare their abilities in solving n-dimensional benchmark optimization problems (Rastrigin function (Eq.~\ref{eq:Rastrigin}), Rosenbrock function (Eq.~\ref{eq:Rosenbrock}), Sphere function (Eq.~\ref{eq:Sphere})). The Rastrigin and Rosenbrock functions are difficult problems due to the existence of multiple local minima in the vicinity of the global minimum, whereas the sphere function has a unique minimum at 0, with smooth gradients towards it. The Rosenbrock function has relatively smooth gradients towards broad local minima region, but it can be difficult to find the global minimum among them. 

\begin{align}
  f_{\textrm{Rastrigin}}(\mathbf{x}) &= An + \sum_{i=1}^{n} \left [x_i^2 - A\cos(2 \pi x_i)  \right ] \label{eq:Rastrigin}\\
  f_{\textrm{Rosenbrock}}(\mathbf{x}) &= \sum_{i=1}^{n-1} \left [100(x_{i+1} - x_i^2) + (1 - x_i)^2 \right ] \label{eq:Rosenbrock}\\
  f_{\textrm{Sphere}}(\mathbf{x}) &= \sum_{i=1}^{n} x_i^2 \label{eq:Sphere}
\end{align}

\begin{figure}
\centering
\includegraphics[width= 0.95\columnwidth]{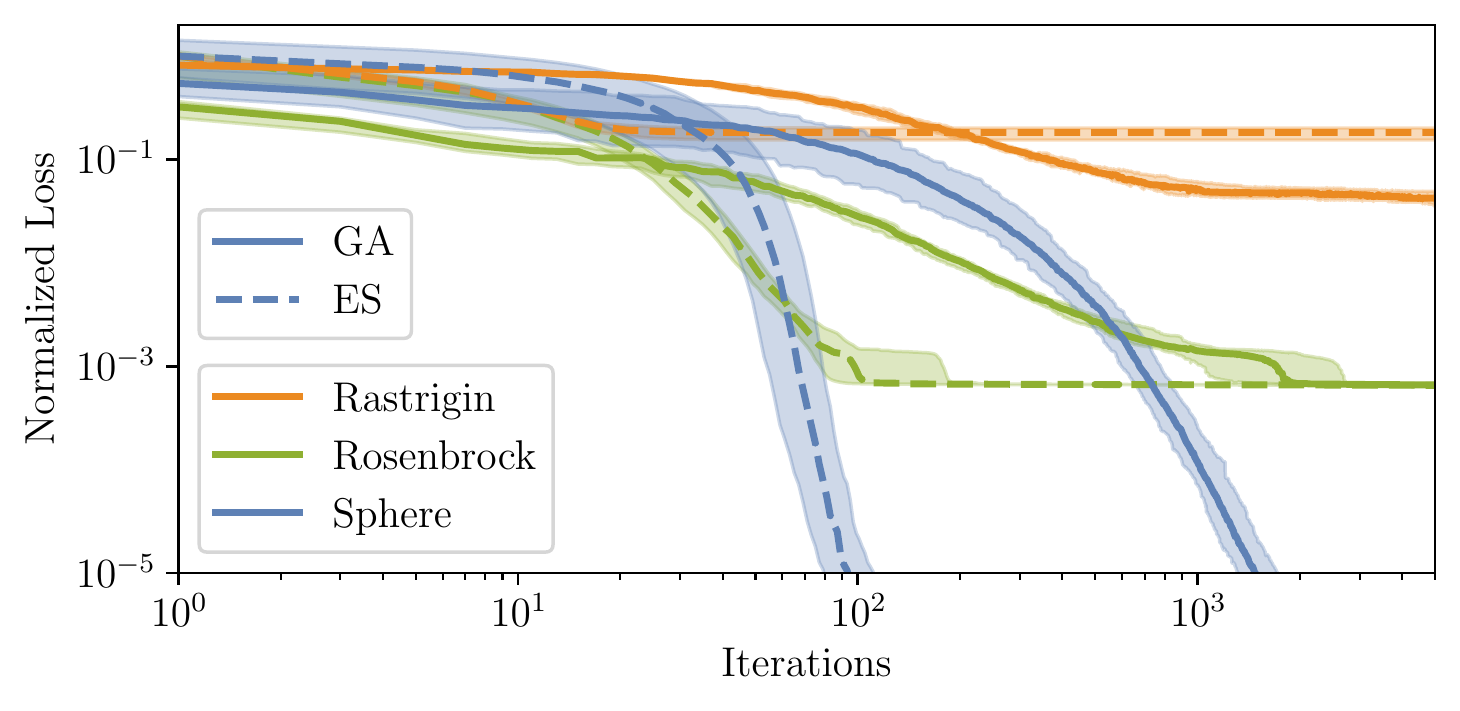}
\vspace{-1.5em}
\caption{Comparisons of the loss dynamics of the ES and GA on the Rastrigin, Rosenbrock, and Sphere functions for $n=50$. The ES can take advantage of functions that have smooth gradients leading to its optima (Sphere, Rosenbrock), whereas the GA can overtake the ES for functions with multiple local optima (Rastrigin). Both algorithms failed to discover the global optima of the Rosenbrock function, and instead found the easier to find local minimum. The loss is normalized to start at 1 by dividing by the medians across independent runs, of the maximum value obtained throughout its evolution. The 25th to the 75th percentiles across 25 independent runs are shaded.}
\label{fig:EA_comparison}
\end{figure}

To avoid any inherent biases either algorithm might have towards solutions of a particular distribution, we translate the loss function (Eq. \ref{eq:Rastrigin}--\ref{eq:Sphere}) in space by a random Gaussian vector. 
To this end we can write the benchmark functions as a function of a new variable $\mathbf{z}$, where $z_i = x_i + c_i$ where $c_i$ are constants sampled from a $\mathcal{N}(0, 1)$ normal distribution, which results in randomization of the optimum's location.

Fig.~\ref{fig:EA_comparison} compares the performance of the 
two algorithms. 
Due to the ES's approximation of the gradient, it is generally able to find the minimum of a smooth loss function faster than the GA (e.g.\ for Sphere function, see Fig.~\ref{fig:EA_comparison}). 
However, due to a fixed variance sampling, it can get stuck in local solution as shown in the Rastrigin function, where the GA surpasses ES.
Also for the Rosenbrock function, both algorithms only converge to the local minima of $\approx 10^{-3}$. 
For both the Rosenbrock and Sphere, the GA is orders of magnitude slower in finding comparable solutions.

Taking into account the significant differences of these two evolutionary algorithms, we strengthen the evidence for the universal utility of criticality for problem solving. 

\subsection{Generalizability}

\begin{figure}
\centering
\includegraphics[width= 0.8\textwidth]{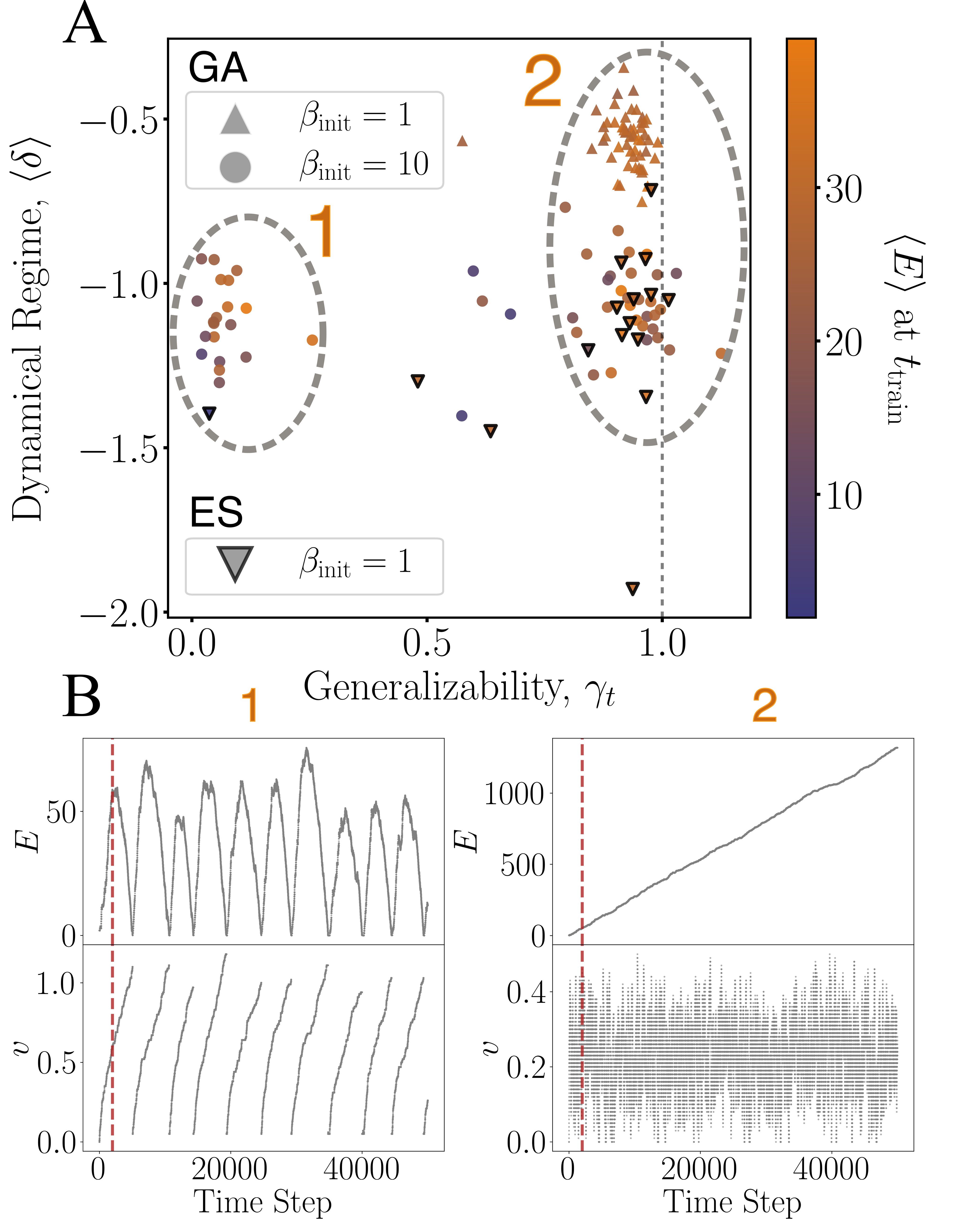}
\vspace{-.5em}
\caption{Populations initialized at criticality find solutions that generalize beyond their training condition, whereas subcritical populations often overfit and fail when organisms lifespan is changed.
{\bf A:} 54 populations of each type (triangles - initially critical, circles - initially subcritical).
After 4000 generations of GA all critical populations reach a high fitness (indicated by colour) and nearly perfect (with one exception) generalizability $\gamma_t$ (Eq.~\ref{eq:generalizability_param}).
The initially subcritical organisms split between badly generalizing cluster 1 (19 populations), generalizable cluster 2 (28 populations), and 7 populations not assigned to any cluster. 
This split is not predicted by their attained fitness.
 For the ES (indicated by the triangles pointing down with black border) the variability of initially critical solutions is larger (13 out of 16 are in cluster 2, one in cluster 1 and two in-between), and initially subcritical organisms $\beta_{\mathrm{init}} = 10$ do not evolve to any reasonable fitness. 
{\bf B:} Energy and velocity as a function of time for representative examples of the organisms from cluster 1 and clusters 2 (marked in panel A). 
The dashed orange line denotes the training lifespan $t_{\mathrm{train}} = 2000$.}
\label{fig:generalizability}
\end{figure}

\fs{ For successful biological systems robustness against environmental change is the paramount feature, therefore, it can be used to determine the success of evolved artificial organisms.}
We propose a simple measure to investigate how the model behaves outside of its explicit training conditions.
Specifically, for a population trained with the the organism's lifespan parameter set at $t_{\mathrm{train}}$ we define generalizability as the speed of growth of the average fitness  if the organism's lifespan is extended to $t_{\mathrm{extend}}$. Formally:
\begin{eqnarray}
\gamma_t = \frac{\langle E_{t_\mathrm{train}} \rangle / t_\mathrm{train}}{\langle E_{t_\mathrm{extend}} \rangle / t_\mathrm{extend}}.
\label{eq:generalizability_param}
\end{eqnarray}
The \emph{stable generalizability}, $\gamma_t = 1$ corresponds to linear growth whereas sublinear behaviour $\gamma_t \ll 1$  indicates possible overfitting to the particular organism's lifespan $t_{\mathrm{train}}$.

\fs{We consider initially critical ($\delta\approx0$, $\beta_{\mathrm{init}} = 1$) and initially subcritical ($\delta\approx-1$, $\beta_{\mathrm{init}} = 10$) populations evolved for 4000 generations and then test their performance for an extended lifespan of $t_\mathrm{extend} = 50000$ time steps (instead of the $t_{\mathrm{train}} = 2000$).}
 As reported in previous sections, the critical populations evolved with GA converge to $\delta \approx -0.41$, and they all have a similar fitness after training.
Interestingly, when increasing the organisms' lifespan, the fitness of the critical population continues to grow linearly, signifying almost perfect generalizability.
About half of the subcritical populations reach the same fitness level.
However, the subcritical populations split up into two clusters: cluster one with generalizability close to 0 and cluster two with generalizability close to 1 (Figure~\ref{fig:generalizability}A).
Surprisingly, there is no difference in fitness between these two clusters.
 We also evolve the same populations with the ES algorithm and obtain a similar picture for the critical population: almost all populations attain generalizability close to 1. The initially subcritical populations ($\beta_{\mathrm{init}} = 10$), however, fail to evolve to a compatible fitness, and therefore we exclude them from the figure.  Out of 16 populations, one did not generalize at all (in cluster 1) but it also did not reach a similar fitness level ($\langle E \rangle_{t_{\mathrm{train}}}  \approx 20$) and two reach an intermediate state. Notably, the 3 populations that did not generalize for the ES were some of the most subcritical solutions found. 
 
 To understand the difference we look more precisely at the both clusters, Figure~\ref{fig:generalizability}B. 
Organisms in cluster 1 reach their maximal fitness/average lifetime energy (sometimes higher than in clusters 2) at the end of their lives, but often quickly lose fitness when tested beyond their training environment. Its velocity profile $v$ offers an explanation to the bad generalization. 
The organisms from cluster 1 follow the strategy to  increase the velocity permanently until the end of their training lifespan (Figure~\ref{fig:generalizability}B).
However, moving with such a high velocity is not compatible with the energy influx from feeding, and they break down shortly after the end of their training lifespan, this demonstrates that these organisms overfit the training conditions.
 The generalizable populations (cluster 2) have a much more complex velocity profile that accelerates and decelerates often, in contrast to its fitness, which grows consistently and linearly beyond the lifespan of their training environment. 

\fs{Overall, the initialization in the critical regime results in almost perfect generalizability of evolved populations, whereas initially strongly subcritical populations risk overfit their training conditions.}

\subsection{Effect of genetic perturbations on the fitness}
\begin{figure}\centering
\includegraphics[width= 0.95\columnwidth]{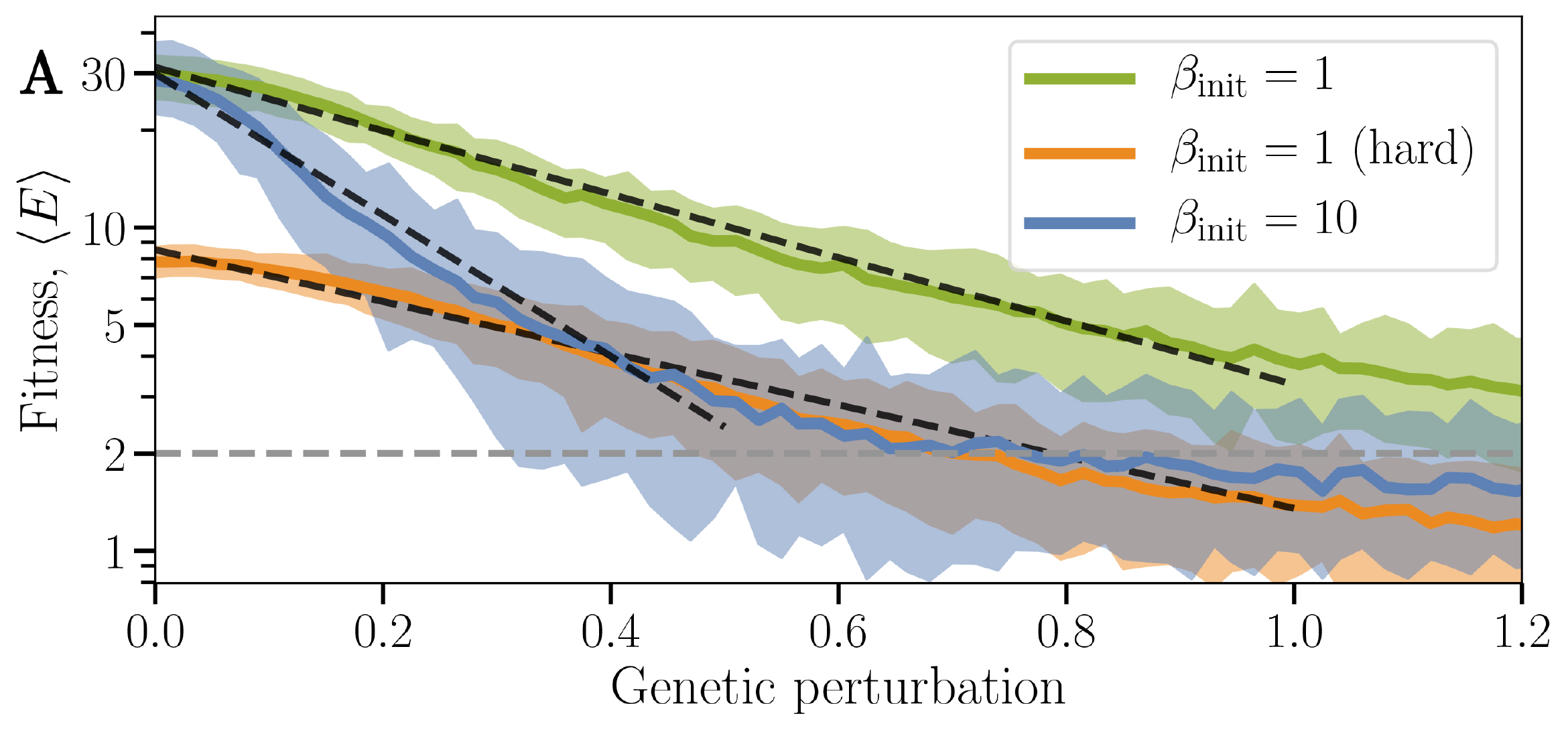}
\vspace{-1em}
\includegraphics[width= .95\columnwidth]{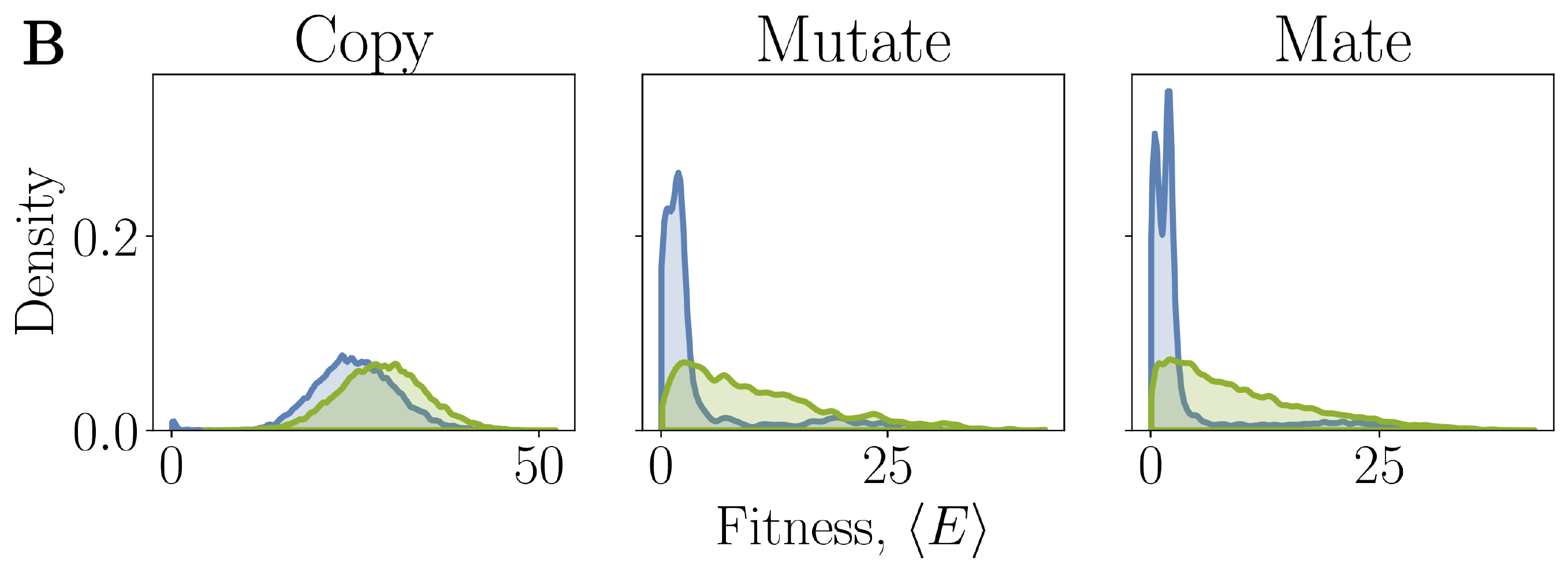}
\caption{Initially critical populations show larger genotypic stability than initially subcritical ones.
\textbf{A:} The phenotype (fitness) as a function of the genotypic perturbation (changes in connectivity) for two initial conditions: $\beta_{\mathrm{init}} = 1$ and $\beta_{\mathrm{init}} = 10$. For the initially critical model we also plot the change in fitness for the hard task (orange line).
Dashed black lines indicate exponential fit with exponent -2.26 (critical), -1.84 (critical, hard) and -5.03 (subcritical).
\textbf{B:} The histograms of the fitness values for nearly fully evolved agents (between generation 3500 and 4000), categorized according to the last evolutionary operator (copy, mutate, or mate) that was applied to them.
The $\beta_{\mathrm{init}} = 10$ agents are less likely to remain fit when their genotype is changed by mutation or mating.}
\label{fig:phenoty_genotype}
\end{figure}

\fs{Next, we examine the stability of the evolved organisms to genetic perturbations.}
We apply genetic perturbations of different magnitudes to the evolved organisms of initially critical $\beta_{\mathrm{init}}=1$ and subcritical $\beta_{\mathrm{init}}$ populations.
We perturb all weights of the connectivity matrix by randomly adding or subtracting a number $f_\mathrm{pert}$ and then evaluate the fitness of the resulting organism.
As expected, we find that fitness decays rapidly with perturbation magnitude for both batches of populations, however the subcritical ones decay much faster (Figure~\ref{fig:phenoty_genotype}A). By characterizing this fitness decay, we can understand better the smoothness of the genotype-phenotype map of these optimized solutions, and how different solutions can be found with varying degrees of robustness.
We evaluate the fitness decay by the slope of an exponential function fitted to the fitness.

 For the hard task, only the fitness decay of the initially critical populations can be evaluated ($\alpha=-1.84$), as the subcritical population was unable to solve this task. For the simple task, $\alpha = -2.26$ and  $\alpha = -5.03$ for the initially critical and subcritical populations, respectively. The subcritical population decays in fitness at more than double the rate than the critical population, indicating a much higher sensitivity of subcritical systems to perturbations. Interestingly, the fitness decay of both critical populations in the easy and hard task are similar.

\fs{The EA is a source of constant genetic perturbations that are necessary in the beginning of evolution but can become detrimental later}.
We consider the individual effect of the evolutionary operators (copy, mutate, and mate) on the resulting fitness of the organisms.
The variability of fitness for copying simply reflects the natural variability in community fitness rankings and organism behaviour.
However, both mating and mutation in fully evolved subcritical populations typically results in a fitness close to 2 -- signifying totally unfit organisms (Figure~\ref{fig:phenoty_genotype}B).
At the same time, initially critical organisms retain diverse fitness values after mutation and mating, some being close to the optimum. This phenotypic diversity allows the originally critical populations to retain their evolvability as opposed to the rigid search performed by strongly subcritical populations which have more discontinuous genotype-phenotype landscapes.

\section{Discussion}

\fs{We demonstrate that in various scenarios evolving populations of agents converge to a moderately subcritical state with the resulting deviation from criticality depending on the task's difficulty. }
This might appear to be a contradiction to the previous studies, suggesting that operating close to criticality is optimal for natural systems \citep{mora2011biological_poised_at_critic, munoz2018colloquium, roli2018dynamical}.
However, a recent body of research showed that for simple tasks, operating at some distance to criticality might be an optimal solution for the sensitivity/stability tradeoff \citep{hidalgo2014information,tomen2014marginally,villegas2016noise_subcritical,cramer_control_2020}.

We observe that the distance from criticality affects an agent's ability to solve complex tasks and to robustly evolve generalizable behaviour, validated using two different evolutionary approaches: a genetic algorithms and an evolutionary strategy.
Specifically, we observe that \textit{slightly subcritical} populations are evolvable for different complexities of the control network and task, whereas \textit{strongly subcritical} populations fail in both algorithms. Interestingly, the evolutionary strategy is more successful in optimizing strongly supercritical populations but less so for strongly subcritical ones.
Given that solving complex tasks and being adaptive are crucial in natural environments, we propose that living systems operate in the subcritical regime in close proximity to the critical point.
Moreover, we show that the optimal regime moves closer to criticality as we increase the task difficulty, which suggests that the optimal distance from criticality varies.
Those findings are confirmed by \citet{cramer_control_2020} as well as \citet{villegas2016noise_subcritical}, who showed that the optimal distance from criticality in the subcritical regime decreases for higher task complexity or larger system size.

We further observe that populations can only become more subcritical during evolution corresponding to a more ordered phase. 
They fail to become more critical (more disordered) even when this would have eventually led to superior behaviour, a phenomena that we suspected to be an artifact of the genetic algorithm used in \citep{Prosi}. However, this does not seem to be the case as confirmed here by using the evolutionary strategy, which replicates this phenomena.
As it is \emph{a priori} unknown which distance from criticality will be optimal when evolving for a new task, initializing at the critical point could be the only way for the evolutionary process to descend to the optimal regime.
However, in the long run this would require some sub-populations to always maintain closeness to criticality.
How this can be achieved for neuronal networks is a subject of vivid research (for a review, see~\citep{zeraati2020self,buendia2020feedback,kinouchi2020mechanisms}) and for the embodied Ising agents it remains open for further investigation.
Maintaining evolvability in simpler systems was, for instance, achieved by switching between different rough energy landscapes \citep{wang2019evolving}.
The inhomogeneity of the environment and coevolution can also contribute to the preservation of the critical regime~\citep{hidalgo2014information}.
 Overall, the maintenance of evolvability throughout evolution is an important question beyond the embodied Ising agents studied here.

Our results extend and partly revise the earlier findings of \citet{khajehabdollahi2020sinas_paper}, that reported a superior evolvability of critical populations and an approximate convergence to criticality during evolution.
We confirm that the critical regime allows reliable evolvability, additionally, we extend our understanding by considering a set of tasks and architectures in the model.
However, our more precise procedure to infer the dynamical regime and the fine sampling of initial conditions uncover additional complex dynamics.
Namely, that critical populations of Ising-agents converge to the subcritical regime, and the distance to criticality depends on the task complexity.
We extend our earlier work \citep{Prosi} and verify that the results are not contingent on the specifics of our Genetic Algorithm (GA) by comparing them to results generated by an Evolution Strategies (ES), an instance of a different family of evolutionary algorithms. 
We confirm that our major findings hold under both optimization algorithms. The ES, however, shows stronger sensitivity to its hyper-parameters, but optimizes the fitness faster. A future study could potentially compare these results with an algorithm with an adaptive $\sigma$ such as Parameter Exploring Policy Gradients \citep{Sehnke2010:PGPE} or 
Covariance Matrix Adaptation Evolution Strategy (CMA-ES).

We also propose a new way to investigate capabilities of the resulting organisms by defining generalizability and genetic stability measures.
Both measures reveal the benefits of staying close to the critical state beyond a simple fitness comparison.

\section{Acknowledgements}

This work was supported by a Sofja Kovalevskaja Award from the Alexander von Humboldt Foundation. SK and EG thank the International Max Planck Research School for Intelligent Systems (IMPRS-IS) for support. G.M., A.L. are members of the Machine Learning Cluster of Excellence, funded by the Deutsche Forschungsgemeinschaft (DFG, German Research Foundation) under Germany’s Excellence Strategy (EXC no. 2064/1) project no. 390727645. We acknowledge the  support from the BMBF through the T\"ubingen AI Center (FKZ: 01IS18039A)

\footnotesize
\printbibliography
\newpage

\appendix

\section{Effects of thermalization time}\label{sec:app:thermal}
\begin{figure}\centering
\includegraphics[width= 0.95\columnwidth]{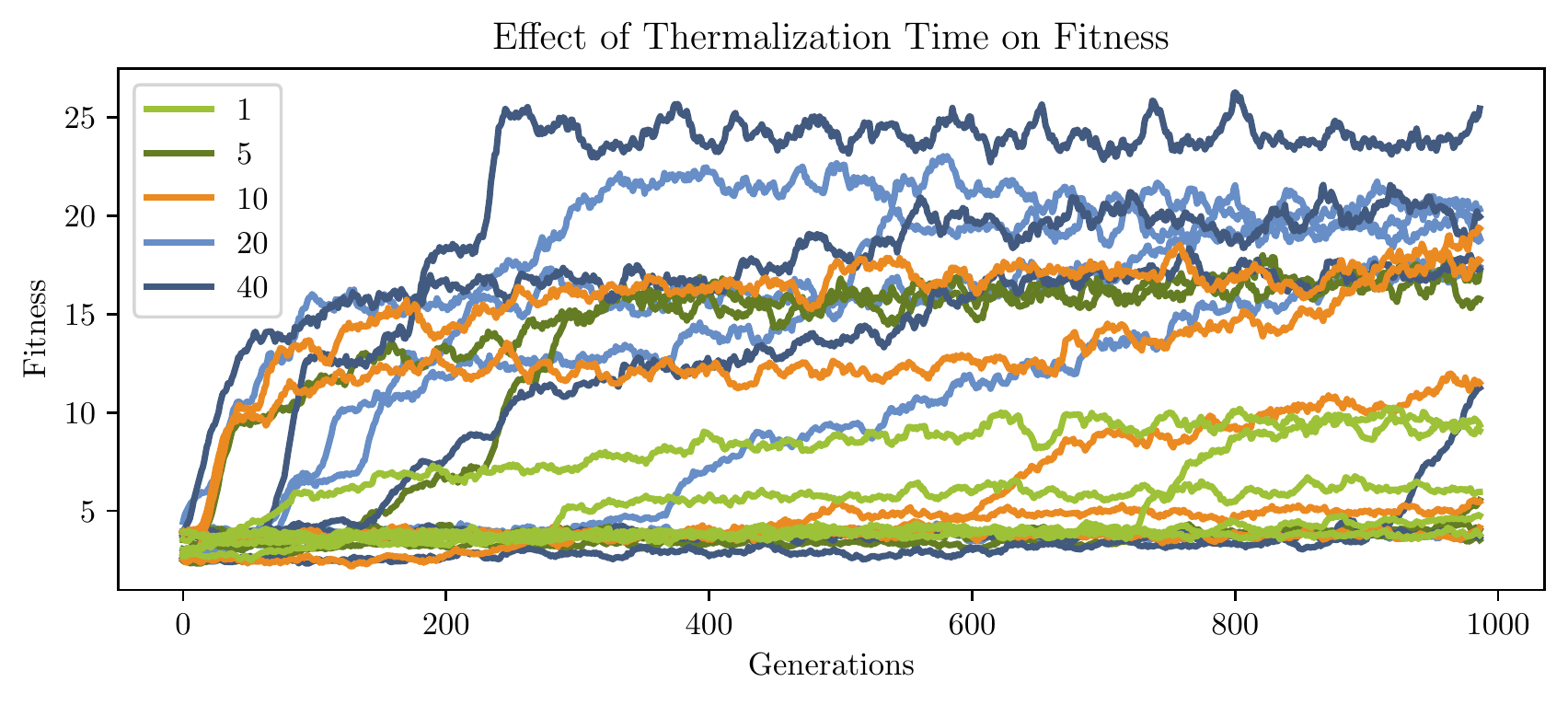}

\caption{The effect of different thermalization times (1, 5, 10, 20, and 40 thermalization steps) is tested for 5 independent populations each, evolved using the GA in the simple task.}
\label{fig:thermalization}
\end{figure}

The Ising networks have a time span during which the system can adapt to the new inputs -- the thermalization time. In Figure~\ref{fig:thermalization} we analyze the dependency of the fitness of evolved individuals on this parameter. We find that the value of 20 yields optimal fitness, however, we have chosen the value of 10 for computational reasons as it provides a good compromise between computational performance and achieved fitness in a given number of generations. Future investigations can be directed to uncover how the thermalization time influences the optimal dynamic regime after convergence or before the beginning of the evolution. 

\section{Distribution of distances to criticality}\label{sec:app:delta_distribution}
In Section \ref{subsec:evo_delta} we discuss the evolution of the dynamical regime as measured by the distance to criticality, summarized by the parameter $\delta$. Specifically, we compare the final values of $\delta$ after 4000 generations of evolution, for both the ES and the GA, for the simple task and the hard task. A box plot and histogram of the data from the final generations of these simulations are plotted here. For each independent simulation, the top 30 most fit agents are selected and their $\delta$ values averaged. We have 54 simulations using the GA and 16 using the ES. (In Figure \ref{fig:simple_vs_hard_task}, the lines are generated from a dataset with 10 and 16 simulations for the GA and ES, respectively. Here we use the larger dataset of 54 simulations for the GA, which only has $\delta$ calculated for its final generation.) A Mann-Whitney U test is used to check if the $\delta$ values of the hard task are higher than the simple task. For both the GA and ES, the test shows the hard task has larger $\delta$s than the simple task, with $p = 2.2 \times 10^{-6}$ and $p = 4.4 \times 10^{-3}$, respectively.

\begin{figure}\centering
\includegraphics[width= 0.95\columnwidth]{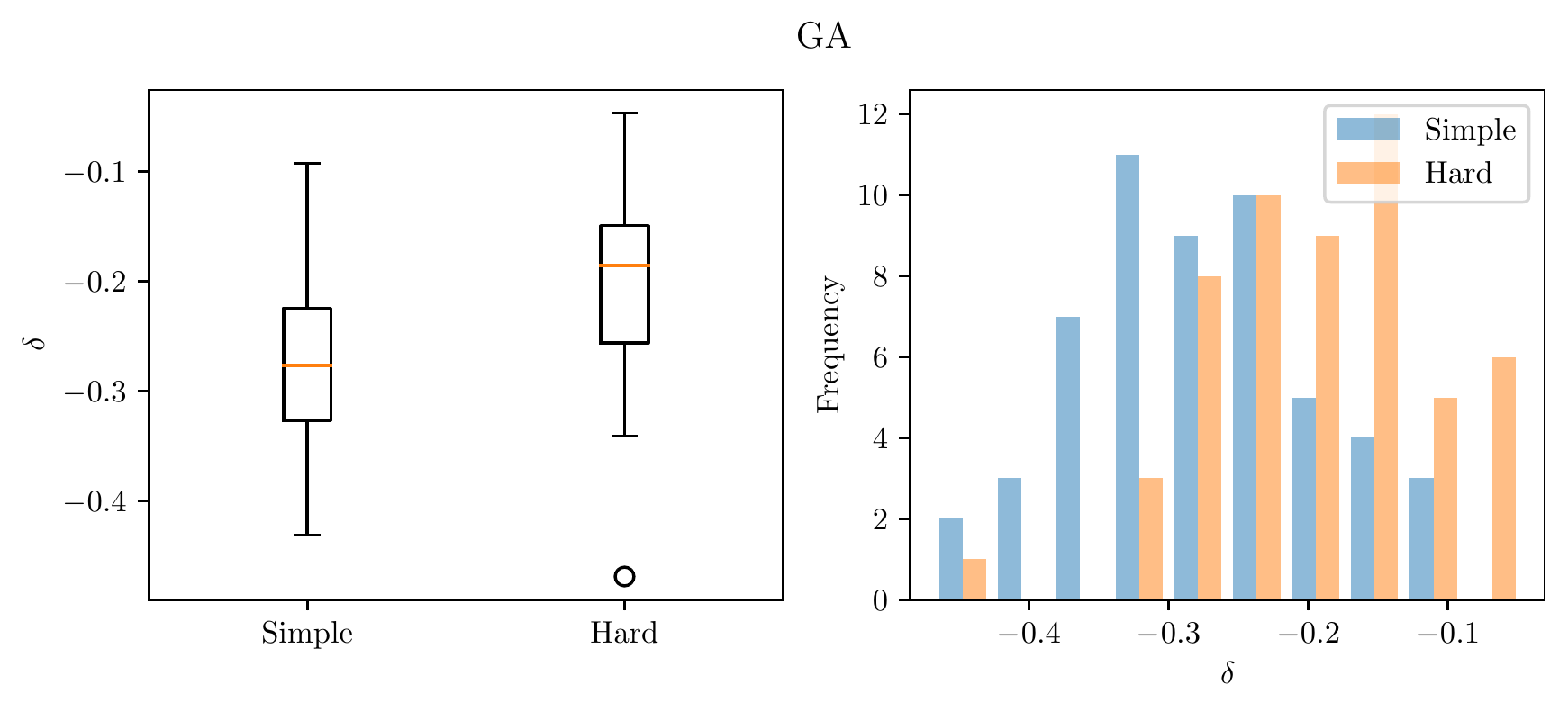}
\caption{Distribution of the final $\delta$ values for 54 independent simulations using the GA. Each simulation has a population of 50, of which only the top 30 most fit individuals have their $\delta$s recorded and then averaged. The hard task has larger $\delta$ values than the simple task with $p = 2.2 \times 10^{-6}$ according to a Mann-Whitney U test.}
\label{fig:delta_dist_GA}
\end{figure}

\begin{figure}\centering
\includegraphics[width= 0.95\columnwidth]{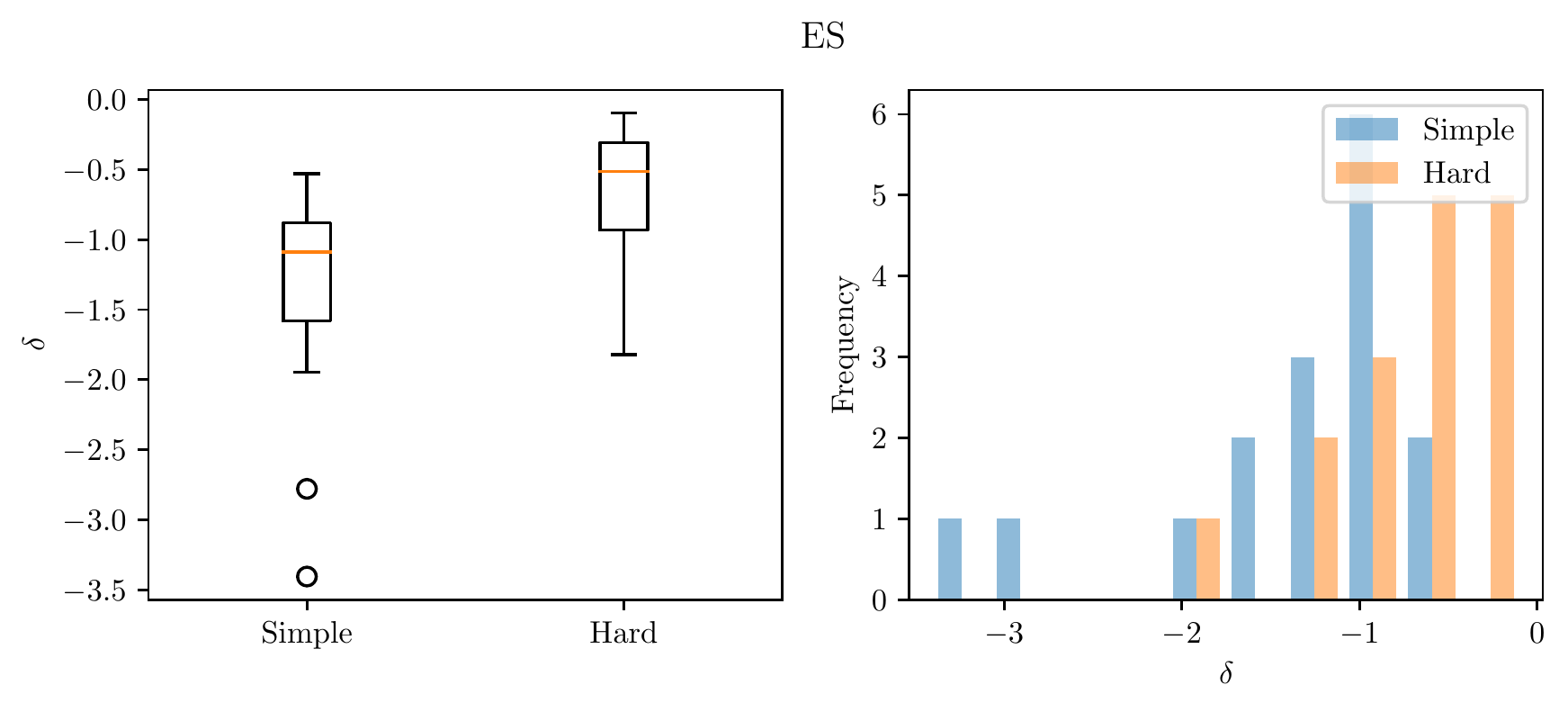}
\caption{Distribution of the final $\delta$ values for 16 independent simulations using the ES. Each simulation has a population of 50, of which only the top 30 most fit individuals have their $\delta$s recorded and then averaged. The hard task has larger $\delta$ values than the simple task with $p = 4.4 \times 10^{-3}$ according to a Mann-Whitney U test.}
\label{fig:delta_dist_ES}
\end{figure}

\end{document}